\documentclass{article}

\usepackage{arxiv}
\usepackage{cite}
\usepackage[utf8]{inputenc} 
\usepackage[T1]{fontenc}    
\usepackage{hyperref}       
\usepackage{url}            
\usepackage{booktabs}       
\usepackage{amsfonts}       
\usepackage{nicefrac}       
\usepackage{microtype}      
\usepackage{amsmath}        
\usepackage{cleveref}       
\usepackage{lipsum}         
\usepackage{graphicx}
\usepackage[numbers,square]{natbib}
\usepackage{doi}

\usepackage{booktabs}
\usepackage{tabularx}
\usepackage{multirow}
\usepackage{booktabs}
\usepackage{graphicx} 

\title{ECG-biometrics-bench: A Unified Framework for Reproducible Benchmarking of ECG Biometrics}

\date{}

\newif\ifuniqueAffiliation
\uniqueAffiliationtrue

\ifuniqueAffiliation 
\author{ \href{https://orcid.org/0000-0002-0804-2040}{\includegraphics[scale=0.06]{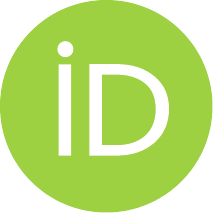}\hspace{1mm}Milad Parvan}\\
	Independent Researcher\\
	Milan, Italy \\
	\texttt{miladparvan72@gmail.com} \\
}

\hypersetup{
pdftitle={ECG-biometrics-bench: A Unified Framework for Reproducible Benchmarking of ECG Biometrics},
pdfsubject={},
pdfauthor={Milad Parvan},
pdfkeywords={First keyword, Second keyword, More},
}

\begin{document}
\maketitle

\begin{abstract}
Electrocardiogram (ECG) biometrics have emerged as a promising modality for continuous, liveness-aware authentication in wearable systems. However, many prior studies report overly optimistic results due to data leakage (e.g., random splits within the same session). To address this issue, we introduce \textit{ECG-biometrics-bench}, a modular, reproducible benchmarking framework that standardizes preprocessing, segmentation, and evaluation across seven widely used public ECG datasets spanning clinical, ambulatory, and large-scale cohort settings. The framework supports both closed-set and open-set (i.e., subject-disjoint generalization in this work) evaluation, as well as progressively realistic protocols including cross-session and long-term temporal separation. To facilitate reproducible research in the community, the \textit{ECG-biometrics-bench} repository will be made publicly accessible on GitHub upon the acceptance of this manuscript. Through a comprehensive multi-dataset analysis, we expose the \textit{Random Split Fallacy}, demonstrating that intra-session evaluation protocols artificially inflate performance while masking severe degradation caused by temporal drift and unseen identities. Furthermore, by evaluating multiple architectures, including DeepECG, ResNet1D, and CNN-LSTM, we show that these failures are not model-specific but are likely inherent to current supervised feature-learning paradigms. Finally, we demonstrate that performance degradation due to temporal aging can be partially mitigated through a \textit{heavy enrollment, lightweight authentication} strategy based on dynamic multi-session template fusion. These findings establish a more realistic baseline for ECG biometrics and highlight critical challenges that must be addressed for reliable real-world deployment.
\end{abstract}


\section{Introduction}

Biometric recognition systems aim to identify individuals based on physiological or behavioral characteristics. These systems are increasingly deployed as an alternative to traditional authentication mechanisms such as passwords and physical tokens. Biometric traits include facial recognition, fingerprinting, iris recognition, and electrocardiogram (ECG) signals, among others. Each biometric trait has its own advantages and disadvantages, which can be selected based on the application at hand. Facial recognition and fingerprint recognition remain the most widely deployed biometric modalities. However, despite their widespread adoption, they remain susceptible to advanced spoofing and presentation attacks, which can pose potential privacy and security challenges. Consequently, research focuses on alternative traits, such as ECG, which is the electrical activity of the heart. ECG offers two key advantages compared to other biometric traits. First, ECG enables continuous or periodic authentication, allowing a system to verify the identity of a user throughout an active session rather than only at login time. Second, ECG signals inherently provide a natural liveness detection mechanism, as they originate from the electrical activity of the heart, presenting a significantly higher barrier against conventional spoofing and replay attacks \cite{rai2025lightweight}. These properties allow ECG signals to extend beyond their traditional medical applications and emerge as a promising biometric modality.

Although ECG biometrics has great potential, it is being developed at a slower pace than other physiological biometric modalities. Firstly, there is a notable scarcity of publicly available ECG datasets suitable for biometric recognition, in contrast to the extensive databases available for other modalities. Moreover, the available public datasets differ significantly in their acquisition settings. The scarcity and heterogeneity of ECG datasets represent a major barrier to standardized evaluation in the field. Secondly, obtaining high-quality ECG signals requires specialized sensors and controlled environments, whereas modalities like fingerprint or face can be captured with standard imaging devices, such as a simple phone camera. Generally, there are two types of ECG recording hardware, namely on-the-person and off-the-person. On-the-person sensors, such as wet electrodes or wearables, typically yield high-quality signals with minimal noise artifacts. However, their intrusive nature and requirement for physical attachment often impede user acceptance. Conversely, off-the-person settings offer greater usability but may be susceptible to environmental noise and signal attenuation \cite{pereira2023biometric}. Perhaps the most critical challenge is the intrinsic temporal variability of ECG signals. ECG morphology can vary due to factors such as stress, physical activity, electrode placement, and changes in health condition \cite{schijvenaars2008intraindividual}. These variations introduce temporal inconsistency in biometric templates and significantly degrade recognition performance over time. These degradations are discussed in more detail in the upcoming sections of this paper. 

Despite these challenges, ECG biometric research has advanced significantly in recent years, particularly with the adoption of deep learning methods. However, the field remains highly fragmented due to the lack of standardized experimental protocols and benchmarking practices \cite{merone2017ecg}. Reproducibility is therefore a critical requirement for advancing ECG biometric research. In contrast to mature machine learning domains such as computer vision, where benchmarks like ImageNet provide a common evaluation ground, ECG biometrics lacks a unified framework for reproducible experimentation \cite{melzi2023ecg}. As a result, reported performance across studies is often difficult to compare. These inconsistencies typically arise from several methodological issues, including arbitrary dataset sub-sampling, undocumented preprocessing pipelines, reliance on random intra-session train-test splits, and evaluation restricted to closed-set recognition scenarios. These limitations and inconsistencies are examined in detail in Section~\ref{sec:motivation}.

To address these challenges, we present \textit{ECG-biometrics-bench}, an open-source and modular benchmarking framework designed to standardize the end-to-end evaluation of ECG biometric systems. The framework provides a unified experimental environment that integrates dataset ingestion, signal preprocessing, augmentation, model training, and biometric evaluation protocols. By enforcing consistent experimental conditions across datasets and models, the framework enables reproducible comparisons and reduces the risk of methodological inconsistencies. The framework is publicly available as an open-source repository to facilitate transparent and extensible research in ECG biometrics.

The primary contributions of this work are summarized as follows:

\begin{enumerate}

\item \textbf{A Unified Benchmarking Framework for ECG Biometrics:}
We introduce \textit{ECG-biometrics-bench}, an open-source modular framework that standardizes the complete experimental pipeline for ECG biometric recognition. The framework decomposes the biometric workflow into independent components covering dataset ingestion, preprocessing, augmentation, model training, and evaluation, enabling reproducible experimentation across different research settings.

\item \textbf{Standardized Dataset Integration:}
The framework provides a unified data ingestion layer supporting seven widely used public ECG datasets, including ECG-ID, Heartprint, CYBHi, MIT-BIH, NSRDB, PTB, and PTB-XL. Dedicated dataset loaders normalize heterogeneous file formats, directory structures, and session annotations into a consistent internal representation.

\item \textbf{Rigorous Temporal and Open-Set Evaluation Protocols:}
Moving beyond the flawed single-session paradigms prevalent in the literature, we implement standardized regimes that strictly enforce cross-session temporal aging and subject-disjoint (open-set) testing. This infrastructure allows researchers to definitively measure genuine biometric generalizability and systematically expose vulnerabilities like the Random Split Fallacy.

\item \textbf{Reproducible Baseline Benchmark:}
Using the proposed framework, we conduct a large-scale benchmark across seven public ECG datasets and report baseline performance under multiple evaluation regimes. These results establish a reproducible reference point for future research and highlight the performance gap between random-split evaluations and realistic biometric evaluation protocols.

\end{enumerate}

The remainder of this paper is organized as follows: Section \ref{sec:motivation} provides the motivation for this work by exploring the disparate evaluation settings prevalent in the literature, which necessitate the creation of a standardized benchmark. Section \ref{sec:methodology} details the different modules of the \textit{ECG-biometrics-bench} framework. Section \ref{sec:results} presents a large-scale systematic benchmark across seven public datasets using a baseline 1D-CNN architecture. Moreover, in this section, we analyze the performance collapse observed under realistic conditions and discuss the implications for future biometric system design. Section \ref{sec:limitations_future_work} addresses the limitations of the current study and outlines prospective avenues for future research. Finally, Section \ref{sec:conclusion} concludes the paper.

\section{Motivation}
\label{sec:motivation}

The transition of biometric systems from static anatomical traits to dynamic physiological signals has positioned the ECG as a promising modality for secure biometric recognition. However, despite the rapid advancement of automated recognition algorithms, a significant gap has emerged between the high performance reported in controlled laboratory studies and the robustness required for real-world deployment. This discrepancy largely stems from methodological inconsistencies across studies, including non-standardized data handling, inappropriate evaluation protocols, and limited consideration of scalability. This section highlights these issues and motivates the need for a unified and reproducible benchmarking framework.

\subsection{Inconsistent Data Handling}

A major limitation in existing ECG biometric studies is the inconsistent use of datasets. State-of-the-art methods are frequently evaluated either on private datasets or on arbitrary subsets of publicly available databases. A prevalent practice involves selecting only a subset of subjects from larger datasets, often favoring recordings with minimal noise or excluding subjects with pathological conditions. Such practices hinder meaningful cross-study comparisons, as reported performance may reflect dataset selection rather than methodological improvements. Table~\ref{tab:subject_counts} illustrates this inconsistency by showing examples of subject selections from three widely used ECG datasets: PTB, MIT-BIH, and NSRDB. For instance, the PTB dataset contains recordings from both healthy individuals and patients with various cardiac conditions. Some studies restrict evaluation to the 52 healthy subjects in the dataset, particularly when the focus is not on evaluating robustness to pathological variability. Although filtering subjects may serve the specific objectives of an individual study, this practice fundamentally invalidates direct performance comparisons against models evaluated on the complete, unfiltered dataset. Across the literature, the number of PTB subjects used for evaluation varies widely, ranging from fewer than 20 subjects to 290 subjects (i.e., full dataset). Similar inconsistencies are observed for the MIT-BIH and NSRDB datasets, as summarized in Table~\ref{tab:subject_counts}.

\begin{table}[htbp]
\centering
\caption{Heterogeneity of Subject Selection Across Studies}
\label{tab:subject_counts}
\begin{tabular}{l|cccccc|ccc|ccc}
\hline
\textbf{Dataset Name} & \multicolumn{6}{c|}{\textbf{PTB}} & \multicolumn{3}{c|}{\textbf{MIT-BIH}} & \multicolumn{3}{c}{\textbf{NSRDB}} \\ \hline
Number of Subjects & 52  & 51  & 100  & 20  & 90  & 43  & 47 & 11 & 10 & 10  & 15 & 18 \\ 
& \cite{labati2019deep} & \cite{pereira2013novel} & \cite{ciocoiu2017comparative} & \cite{zhao2011ecg} & \cite{tantawi2015wavelet} & \cite{irvine2008eigenpulse} & \cite{zhang2017heartid} & \cite{jahiruzzaman2015ecg} & \cite{sasikala2010identification} & \cite{camara2018real} & \cite{venkatesh2010human} & \cite{zhang2017heartid} \\
\hline
\end{tabular}
\end{table}

Beyond subject selection, preprocessing pipelines for ECG signals are rarely standardized \cite{mesinovic2025survbench, vest2018open}. Differences in bandpass filtering ranges, normalization strategies (e.g., Z-score versus Min–Max scaling), and R-peak detection algorithms introduce additional sources of variability across studies. As a result, reported performance improvements may originate not from the proposed model architecture but from undisclosed preprocessing steps. This lack of transparency contributes to a broader reproducibility challenge, where replicating reported results requires reconstructing preprocessing pipelines that are insufficiently documented.

\subsection{Inconsistent Train–Test Split (Inter-Session Variability)}

Another widespread issue in ECG biometric research concerns the evaluation protocol used to split training and testing data. Many studies rely on random splitting within the same recording session, where individual heartbeats from a single session are randomly assigned to both the training and testing sets. While this approach simplifies experimentation, it often leads to overly optimistic performance estimates. When samples from the same session appear in both training and testing sets, models may exploit session-specific characteristics or recording artifacts instead of learning subject-invariant features. Consequently, identity recognition effectively becomes a simple classification task rather than a realistic biometric recognition problem. In practical biometric systems, recognition must remain reliable across time, where factors such as electrode placement, recording conditions, and physiological state may vary significantly. These temporal variations introduce a phenomenon commonly referred to as template aging. Evaluation protocols that ignore such variability fail to capture the fundamental challenge of long-term biometric recognition. Empirical evidence demonstrates the impact of this issue. For example, \cite{d2023advancing} reported identification accuracy of approximately 99\% on the CYBHi dataset when training and testing samples were drawn from the same recording session. However, when the evaluation was conducted across separate sessions, performance dropped to approximately 68–70\%. Such findings suggest that many reported results significantly overestimate real-world biometric performance.

\subsection{Inconsistent Train–Test Split (Scalability)}

A scalable biometric system must also demonstrate the ability to generalize to previously unseen individuals. This requirement is typically evaluated through a subject-disjoint protocol, in which the identities used during training are completely different from those used during testing. However, most ECG biometric studies rely on closed-set evaluation protocols, where training and testing are performed on the same set of subjects. While suitable for small-scale applications, such as personal device authentication, this setup does not reveal whether the model has learned generalizable identity representations. In closed-set scenarios, deep neural networks may simply memorize the characteristics of the training subjects rather than learning transferable features. Subject-disjoint (or open-set) evaluation addresses this limitation by separating the identity sets used for training and testing. For example, a model may be trained on Subjects 1–50 and evaluated on Subjects 51–100. This protocol assesses whether the learned embedding space can discriminate between previously unseen individuals without additional training. Although such protocols have become standard practice across more established biometric modalities, they remain relatively uncommon in ECG biometric research.

Scalability is closely linked to the concept of biometric uniqueness, which assumes that physiological signals contain sufficient discriminative information to uniquely identify individuals. While numerous studies claim that ECG signals possess this property, most experiments have been conducted on relatively small datasets. In contrast, modalities such as face and iris recognition benefit from large-scale datasets containing thousands of subjects. Consequently, existing evidence for the uniqueness and scalability of ECG-based biometrics remains limited. Evaluating systems under subject-disjoint conditions provides an alternative way to investigate this issue by assessing whether learned representations remain discriminative as the population size increases. Establishing such evaluation protocols is therefore essential for determining whether ECG biometrics can support large-scale real-world deployments.

\subsection{Reproducible Benchmarking}
There is a growing consensus regarding the urgent need for reproducible, standardized evaluation protocols in ECG biometrics. Recently, \cite{melzi2023ecg} introduced the open-source ECGXtractor framework to benchmark their proposed Autoencoder and Siamese CNN architecture across four datasets. Crucially, their work corroborated the severe degradation caused by temporal aging, demonstrating that EER universally drops when transitioning from intra-session to cross-session evaluations. While their findings validate the existence of this temporal degradation, their framework primarily served to evaluate a specific proprietary topology. Consequently, it remains ambiguous whether this degradation is an artifact of specific models or endemic to the modality itself. Our framework, \textit{ECG-biometrics-bench}, builds upon this foundation by providing a strictly architecturally-agnostic infrastructure. By systematically evaluating diverse models (e.g., DeepECG, ResNet1D, CNN-LSTM) across seven highly heterogeneous datasets, we definitively prove that phenomena like the `Random Split Fallacy' are modality-wide flaws rather than isolated architectural artifacts.

Beyond architectural flexibility, our framework significantly expands the depth of the experimental pipeline compared to existing tools. Specifically, \textit{ECG-biometrics-bench} introduces a standardized Regime Mapping Protocol capable of natively handling various types of datasets. Additionally, it actively facilitates open-set generalizability testing and supports dynamic multi-session template fusion strategies (e.g., the Leave-Last-Out protocol), providing researchers with the infrastructure not just to measure temporal degradation, but to evaluate longitudinal solutions.

In summary, these observations highlight three major methodological limitations in the current ECG biometric literature: (1) inconsistent dataset usage and preprocessing pipelines, which hinder reproducibility and fair comparison; (2) evaluation protocols that rely on session-shuffled train–test splits, which inflate reported performance; and (3) limited assessment of scalability due to the widespread use of closed-set evaluation protocols. Together, these issues create a gap between reported laboratory performance and the requirements of real-world biometric systems. To address these challenges, this study proposes a unified benchmarking framework designed to standardize dataset handling, enforce rigorous evaluation protocols, and facilitate reproducible comparison across ECG biometric models.

\section{Methodology}
\label{sec:methodology}

A biometric recognition system operates by identifying or verifying individuals based on their unique physiological or behavioral characteristics or traits. ECG-based biometric systems employ a structured pipeline that includes data acquisition, preprocessing, feature extraction, matching, and decision-making. Deep learning approaches often integrate feature extraction within the model, thereby eliminating the need for a separate feature extraction stage. ECG signals are typically recorded using electrodes or wearable sensors, and the quality of these signals depends on factors such as sampling rate, electrode placement, and recording conditions. Subsequently, noise and artifacts, such as baseline wander and muscle noise, are removed to improve signal clarity. Common preprocessing techniques include bandpass filtering, signal normalization, and segmentation, which ensure that the input data is consistent and suitable for feature extraction. Features such as temporal intervals, morphological patterns, or learned embeddings from deep learning models are then extracted to represent the individual's ECG signal, capturing the unique aspects of cardiac activity. Biometric recognition tasks are generally categorized as identification or verification. In verification, features from a query signal are compared to a stored template of the claimed individual, while identification involves comparing the query across a database to find the closest match. Similarity between signals is typically measured using metrics such as Euclidean distance, cosine similarity, or correlation coefficients. Based on the similarity score and predefined thresholds, the system makes a decision. In verification tasks, which involve one-to-one comparisons, the system determines whether the query signal belongs to the claimed identity.

The proposed \textit{ECG-biometrics-bench} framework is designed to provide a unified and reproducible benchmarking environment for ECG-based biometric recognition systems. The framework decomposes the biometric pipeline into modular components, including dataset ingestion, preprocessing, augmentation, model training, and evaluation. Each module operates through standardized interfaces, enabling different models, preprocessing strategies, and datasets to be evaluated under consistent experimental conditions. This modular architecture ensures that differences in reported performance arise from algorithmic design rather than inconsistencies in dataset handling or evaluation protocols.

\subsection{Dataset Ingestion and Standardization}
Public ECG datasets vary significantly in terms of file formats, sampling rates, and temporal structure (e.g., continuous 24-hour holters or discrete 10-second records). A primary contribution of this framework is the unified loader API, implemented in the \texttt{load\_dataset} module. This module creates a standardized abstraction layer that logically partitions every dataset, regardless of its native structure, into a uniform evaluation format comprising an explicit enrollment phase and a separate probe phase. The framework integrates seven major public ECG datasets, summarized in Table~\ref{tab:dataset_comparison}.

\begin{itemize}
    \item ECG-ID \cite{lugovaya2005biometric}: The ECG-ID database is a publicly available dataset specifically designed for biometric research, featuring 310 ECG recordings obtained from 90 subjects, with Waveform Database (WFDB) format provided by Physionet \cite{physionet2000}. The data consists of single-lead (Lead I) signals, each recorded for 20 seconds and digitized at 500 Hz. A key characteristic of this dataset is that the number of records per subject varies from 2 to +20, collected over time spans ranging from a single day to six months. 
    
    \item Heartprint \cite{islam2022Heartprint}: The Heartprint database is a comprehensive biometric repository featuring multi-session ECG signals collected over a ten-year period. It contains 1,539 records in text format from 199 healthy subjects (130 males and 69 females), aged 18 to 68, primarily from South Asian and Arabian ethnic groups. The signals were captured at the fingertip level using the dry electrodes of a ReadMyHeart device with a sampling frequency of 250 Hz for 15 seconds per recording.
    
    \item CYBHi (Check Your Biosignals Here) \cite{da2014check}: CYBHi is a public ECG biometrics database designed for research. Unlike clinical databases, it is off-the-person, with signals digitized at 1 kHz and 12-bit resolution. The signals, released as text, resemble Lead I of a standard ECG. The database comprises two subsets: a short-term set (65 participants, two sessions two days apart) and a long-term set (63 participants, two sessions three months apart). While the short-term subset evaluates resilience to immediate day-to-day physiological fluctuations and minor sensor placement variations, the long-term subset rigorously tests the overarching robustness and temporal invariance of biometric systems.
    
    \item MIT-BIH Arrhythmia \cite{moody2001impact}: This dataset is a widely recognized benchmark for ECG analysis, consisting of 48 fully annotated 30-minute two-channel ambulatory ECG recordings from 47 individuals suffering from various forms of arrhythmia. The signals are primarily recorded using modified limb lead II (MLII) and lead V1, digitized at 360 Hz in WFDB format.
    
    \item MIT-BIH Normal Sinus Rhythm (NSRDB) \cite{goldberger2000physiobank}: This dataset is a collection of 18 long-term ECG recordings obtained from 18 subjects (5 men, aged 26 to 45, and 13 women, aged 20 to 50) who were found to have no significant arrhythmias. The recordings are approximately 24 hours long, sampled at 128 Hz in WFDB format, and include two channels.
    
    \item PTB Diagnostic Database \cite{bousseljot1995nutzung}: The PTB Diagnostic ECG Database is a widely used clinical dataset that has also been extensively adopted in ECG biometric research. This dataset contains 549 records from 290 subjects. It features a diverse population, including 52 healthy volunteers and subjects suffering from various heart diseases, primarily myocardial infarction. The signals are recorded using 15 leads (12 standard and 3 Frank leads) and digitized at 1000 Hz.
    
    \item PTB-XL \cite{wagner2020ptb}: The PTB-XL database is a large-scale, publicly available dataset comprising 21,837 clinical ECG records from 18,885 subjects. Captured in the WFDB format, the signals consist of 12-lead ECG data recorded for 10 seconds each, sampled at 500 Hz. The dataset represents a diverse, large-scale biometric scenario, including a wide range of cardiac pathologies and healthy controls annotated by cardiologists.
    
\end{itemize}

\begin{table}[htbp]
\centering
\caption{Summary and Comparison of the ECG Datasets Integrated into the Framework}
\label{tab:dataset_comparison}
\resizebox{\textwidth}{!}{
\begin{tabular}{lccccccl}
\hline
\textbf{Dataset} & \textbf{Subjects} & \textbf{Records} & \textbf{Leads} & \textbf{Freq. (Hz)} & \textbf{Duration} & \textbf{Format} & \textbf{Primary Focus} \\ \hline
ECG-ID & 90 & 310 & 1 & 500 & 20s & WFDB & Biometric Research \\
Heartprint & 199 & 1539 & 1 & 250 & 15s & Text & Long-term Stability (10yr) \\
CYBHi & 63--65 & 250+ & 1 & 1000 & Variable & Text & Off-the-person / State Robustness \\
MIT-BIH & 47 & 48 & 2 & 360 & 30m & WFDB & Arrhythmia / Clinical \\
NSRDB & 18 & 18 & 2 & 128 & 24h & WFDB & Normal Rhythm / Circadian Drift \\
PTB & 290 & 549 & 15 & 1000 & Variable & WFDB & Diagnostic / High-Resolution \\
PTB-XL & 18885 & 21837 & 12 & 500 & 10s & WFDB & Large-Scale / Clinical Diversity \\ \hline
\end{tabular}
}
\end{table}

\subsubsection{The Regime Mapping Protocol}
\label{sec:regime_mapping}

To address the structural heterogeneity of public ECG repositories, which range from short clinical recordings (e.g., PTB-XL) to long continuous Holter monitoring signals (e.g., NSRDB), our framework introduces a \textit{regime mapping protocol}. This abstraction layer decouples the physical organization of the data (files, folders, timestamps) from its logical biometric usage. In practice, publicly available ECG datasets used in biometric research can generally be categorized into three groups. First, some datasets are explicitly structured with clearly defined recording sessions, such as Heartprint and CYBHi, making them naturally suitable for real-world biometric evaluation. Second, several widely used datasets, including collections derived from MIT-BIH databases, consist of long continuous recordings originally designed for clinical analysis but frequently repurposed for biometric studies. Finally, datasets such as ECG-ID, PTB, and PTB-XL do not follow a consistent subject-level structure; they contain varying numbers of recordings per subject with irregular temporal gaps between acquisitions.

These structural differences motivate the need for a standardized evaluation protocol capable of adapting to dataset-specific characteristics while maintaining consistent biometric evaluation criteria. The proposed regime mapping protocol provides this functionality by defining dataset-aware evaluation regimes that align each repository with biometric recognition scenarios. As summarized in Table~\ref{tab:regime_mapping}, the regime mapping protocol standardizes enrollment and probe construction across three categories of ECG repositories.

Heterogeneous multi-record datasets without standardized sessions, such as ECG-ID, PTB Diagnostic, and PTB-XL, contain multiple recordings per subject but do not follow a uniform session structure. The number of recordings per subject and the temporal gap between them vary significantly. For these datasets, the protocol relies on deterministic record-order logic derived from available metadata to construct consistent biometric regimes. Several evaluation regimes are defined to capture different biometric challenges. In the \textit{single-cross-session} regime, the first recording of each subject is used for enrollment, and the second recording is used for probing, enabling cross-record stability evaluation. For short-term intra-day analysis, two complementary regimes are defined. The \textit{single-shot short-term} regime enrolls using the first recording of the first day and tests on the remaining recordings from the same day, representing a minimal enrollment scenario. Conversely, the \textit{leave-last-out short-term} regime constructs the template from all but the last recording of the day and evaluates performance on the final recording, representing a multi-shot enrollment scenario. Longitudinal stability is evaluated through two additional regimes. In the \textit{single-shot long-term} regime, all recordings from the first day form the enrollment template while recordings from subsequent days act as probes. Finally, the \textit{leave-last-out long-term} regime constructs the enrollment template using recordings from past days and evaluates the model on the last available recording day.

Session-structured datasets, such as CYBHi and Heartprint, already provide explicitly defined sessions or acquisition conditions. In these cases, the regime mapping protocol directly leverages the provided session structure to construct biometric evaluations. In the CYBHi dataset, the short-term baseline condition (CI) is evaluated against recordings collected during arousal tasks (A1 and A2), enabling assessment of both short-term stability and physiological-state robustness. Similarly, the long-term CYBHi subset evaluates enrollment on session S1 against probes from session S2. In the Heartprint dataset, enrollment is performed using the first session, while probe data are drawn from later sessions representing different temporal or cognitive conditions, such as reading tasks (S3R) and other longitudinal recordings. These regimes allow systematic evaluation of template aging, cognitive-state variation, and physiological robustness under controlled acquisition conditions.

Continuous long-duration clinical datasets, such as MIT-BIH and NSRDB, consist of single continuous recordings per subject rather than discrete sessions. Since file-based splitting is not applicable in this setting, the framework replaces session-based evaluation with deterministic temporal windowing. In this configuration, enrollment and probe sets are defined as non-overlapping time ranges within the same recording. Unlike random sub-sampling, which may introduce temporal leakage between training and testing samples, this strategy enforces strict temporal separation between enrollment and probe segments. The specific temporal windows are selected according to the experimental objective and may represent short-term stability or long-range temporal variation within the recording. This custom splitting mechanism enables evaluation of intra-record stability and continuous recognition scenarios, where biometric authentication must remain reliable over extended monitoring periods.

By adapting the splitting strategy to the structural characteristics of each dataset category, the regime mapping protocol ensures that biometric performance metrics reflect the intended evaluation challenge, such as short-term stability, physiological robustness, or long-term template aging, rather than artifacts introduced by inconsistent dataset handling procedures.

\begin{table*}[ht]
\centering
\caption{The Regime Mapping Protocol: Standardization of Enrollment and Probe Splits Across Datasets.}
\label{tab:regime_mapping}
\resizebox{\textwidth}{!}{%
\begin{tabular}{@{}lllll@{}}
\toprule
\textbf{Dataset} & \textbf{Regime (Split Mode)} & \textbf{Train / Enrollment} & \textbf{Test / Probe} & \textbf{Biometric Challenge} \\ \midrule

\textbf{ECG-ID}         & single-cross-session      & 1st Record                  & 2nd Record                &  Cross-Record Stability \\
and   & single-shot-short-term    & 1st Record (Day 1)          & Rest of Records (Day 1)   & Short-Term Stability (Single-Shot) \\
\textbf{PTB Diagnostic}           & leave-last-out-short-term & All but Last Record (Day 1) & Last Record (Day 1)       & Short-Term Stability (Multi-Shot) \\
  and          & single-shot-long-term     & All Records (Day 1)         & All Records (Future Days) & Long-Term Stability (Single-Shot) \\
\textbf{PTB-XL}         & leave-last-out-long-term  & All Past Days               & Last Recording Day        & Long-Term Stability (Multi-Shot) \\ \midrule

{\textbf{Heartprint}} 
 & cross-session & session1 & session2  & Long-Term Stability \\
 & cross-session & session1 & session3R & Physiological State Robustness \\
 & cross-session & session1 & session3L & Very Long-Term Stability \\ \midrule

{\textbf{CYBHi}} 
 & cross-session & short-term\_CI & short-term\_A1 / A2 & Short-Term \& Physiological State Robustness \\
 & cross-session & long-term\_S1  & long-term\_S2       & Long-Term Stability  \\ \midrule

{\textbf{MIT-BIH Arr}} 
 & custom-split & A custom range & Another custom range & Intra-Record Stability \& Continuous Recognition\\ \midrule

{\textbf{NSRDB}} 
  & custom-split & A custom range & Another custom range & Intra-Record Stability \& Continuous Recognition\\ \bottomrule
\end{tabular}%
}
\end{table*}

\subsubsection{Dataset Loading and Orchestration}
\label{sec:dataset_loader}

The \textit{ECG-biometrics-bench} framework centralizes dataset handling within the \texttt{load\_dataset} module, which acts as an abstraction layer between raw ECG repositories and the biometric evaluation pipeline. Given the significant heterogeneity of public ECG datasets, the loader provides a unified interface that converts repository-specific formats into a standardized representation suitable for biometric evaluation.

The dataset loading process follows a deterministic orchestration pipeline.

\begin{enumerate}
    \item \textbf{Automated Data Ingestion.}  
    Public ECG repositories are automatically downloaded and extracted when necessary. The loader resolves repository-specific directory structures and organizes recordings into a consistent internal format, ensuring that downstream modules can access signals independently of their original storage layout.

    \item \textbf{Metadata Parsing and Subject-Level Structuring.}  
    Dataset-specific metadata (e.g., subject identifiers, recording timestamps, and session labels) are parsed and mapped into a unified subject-centric structure. This representation allows datasets with heterogeneous organizational schemes to be processed using a common evaluation pipeline.

    \item \textbf{Preprocessing Integration.}  
    Raw ECG signals are optionally resampled and normalized to ensure consistent segment lengths and sampling characteristics across datasets. These preprocessing operations prepare the signals for downstream feature extraction and neural network models. A detailed description of the preprocessing procedures is provided in Section~\ref{subsec:preprocessing}.

    \item \textbf{Regime-Aware Data Splitting.}  
    The loader interfaces directly with the regime mapping protocol described in Section~\ref{sec:regime_mapping}. Based on the structural properties of each dataset, it constructs enrollment and probe sets according to the selected evaluation regime.
\end{enumerate}

By separating dataset-specific parsing from the biometric evaluation pipeline, the \texttt{load\_dataset} module ensures reproducibility and enables consistent benchmarking across heterogeneous ECG repositories.

\subsection{Signal Preprocessing}
\label{subsec:preprocessing}
\begin{figure}
    \centering
    \includegraphics[width=0.75\linewidth]{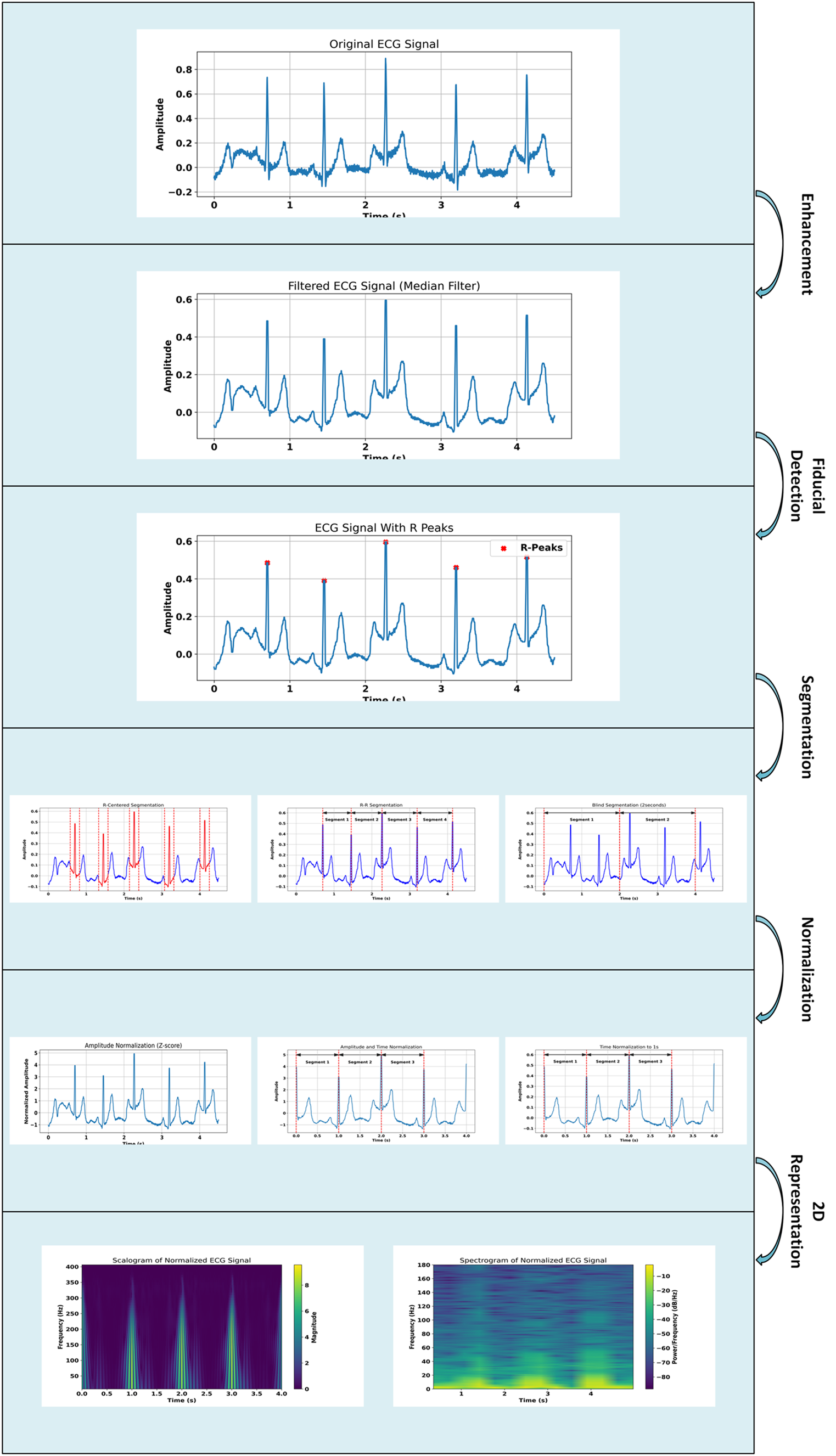}
    \caption{Visual Examples of Preprocessing Methods}
    \label{fig:visual}
\end{figure}

Preprocessing plays a critical role in ECG-based biometric recognition systems by ensuring that the input signals are suitable for subsequent stages of the recognition process. Raw ECG signals are often affected by various noise sources, such as powerline interference, baseline wander, and muscle artifacts, making it challenging to extract meaningful patterns. Therefore, improving the quality of the ECG signal using preprocessing techniques is essential. Common preprocessing approaches involve filtering techniques to remove noise, detecting key fiducial points such as the P-wave or QRS complex, segmenting the signal into relevant intervals, and normalizing the signal. Additionally, some preprocessing techniques have explored alternative signal representations, such as 2D transformations like spectrograms or scalograms, which can capture spatial or frequency-domain features for improved recognition performance. Figure~\ref{fig:visual} shows the different stages of preprocessing visually for a sample real ECG signal.

\subsubsection{Filtering}
Raw ECG signals are susceptible to baseline wander (respiration/movement), powerline interference, and electromyographic (EMG) noise. To preserve biometric morphology while reducing these distortions, several signal enhancement strategies are traditionally employed.
\begin{itemize}
    \item Bandpass Filtering: The most common approach uses frequency-domain filtering, typically between 1 and 40 Hz. For instance, \cite{labati2019deep} utilizes a third-order high-pass Butterworth filter at 0.5 Hz to eliminate baseline wander, often paired with a notch filter for powerline noise. The Butterworth filter is favored for its low passband ripple and rapid roll-off \cite{zehir2024empirical}.
    
    \item Median and Moving Average Filters: These temporal filters target impulse noise and high-frequency fluctuations. Median filters effectively remove high-amplitude artifacts while preserving signal information better than standard frequency filters, often used alongside Discrete Wavelet Transforms (DWT) \cite{sasikala2010identification}. Moving averages offer computational efficiency for smoothing small variations but are frequently combined with other methods, such as Hanning windows, to prevent the loss of critical peaks \cite{belo2020ecg}.
    
    \item Polynomial Fitting: Baseline drift can be mitigated by fitting and subtracting a low-order polynomial. While simple and efficient, it is sensitive to non-linearity. The Savitzky-Golay filter, a polynomial-based smoothing technique, is often used to preserve sharp transitions and higher-order moments \cite{pinto2017towards}.
    
    \item Wavelet Transforms: DWT-based methods decompose signals into frequency components to suppress non-stationary noise like EMG interference. Common implementations include third-scale DWT \cite{hejazi2016ecg} and soft thresholding \cite{el2022wavelet}.
    
    \item Deep Learning Denoising: Recent advancements utilize adaptive reconstruction, such as Conditional Generative Adversarial Networks (CGAN). Specifically, CAE-CGAN architectures employ convolutional autoencoders to maintain spatial locality and latent feature representations while denoising complex signal artifacts \cite{wang2022ecg}.
\end{itemize}

The \textit{ECG-biometrics-bench} framework provides a comprehensive filtering class that supports various techniques, including Butterworth IIR, FIR, Notch, Savitzky-Golay, and DWT-based denoising, allowing users to modularly swap methods within their preprocessing pipelines.

\subsubsection{Segmentation}

The duration of ECG signals used for enrollment and testing is important for designing ECG-based biometric systems. It affects the balance between recognition accuracy, computational efficiency, and user experience. Additionally, it is essential to apply the same procedures for both enrollment and testing data, as the ECG signal can begin at any point within an ECG beat, which can lead to incompatibilities between the two sets of data and even between samples of each set. While various segmentation strategies have been explored in the literature, there is currently no universally accepted optimal duration. Traditional research emphasizes the physiological importance of the P-wave, QRS complex, and T-wave, noting that accurate identification of these components enhances recognition systems. While various computer-based techniques exist for R-peak detection, ranging from wavelet transforms and the classic Pan-Tompkins algorithm to modern deep learning-based CNN detectors, their performance remains sensitive to signal noise and morphological variability. There are generally two main approaches to segmenting ECG signals: fiducial point-based segmentation and blind segmentation.

\begin{itemize}
    \item Fiducial (Beat-Based) Segmentation: This approach relies on the precise localization of the fiducial points (e.g., R-peaks) to align segments with the cardiac cycle. Some studies employ R-centered segmentation \cite{al2024person}, while others use different sample portions before and after the R-peak, such as 200 ms before and 400 ms after \cite{el2022wavelet}. To ensure that no part of the ECG beat is omitted, some studies segment entire intervals, such as R-R or P-P intervals, capturing the whole beat \cite{alduwaile2021using}. Many open-source libraries have been created to help automatically detect the QRS complex in ECG signals. Our framework uses NeuroKit2 \cite{makowski2021neurokit2}, which offers tools for processing physiological signals, including methods for detecting R-peaks using algorithms like Pan-Tompkins \cite{pan1985real}, Hamilton \cite{hamilton2002open}, and Christov \cite{christov2004real}.

    \item Blind (Non-Fiducial) Segmentation: To bypass the fragility of peak detection, this approach uses fixed-length sliding windows (e.g., 5 seconds). This method captures the inter-beat interval and Heart Rate Variability (HRV) context, making it inherently more robust against noise-induced detection failures and arrhythmias. Furthermore, it facilitates data augmentation through overlapping window shifts \cite{jyotishi2021ecg}.
\end{itemize}

The \textit{ECG-biometrics-bench} framework supports both segmentation strategies, allowing for a comparison between classical and modern approaches. The framework supports integration with major open-source R-peak detectors and provides a modular interface for users to experiment with various window lengths and overlaps.

\subsubsection{2D Representations}
Transforming one-dimensional ECG signals into 2D representations is a preprocessing step that allows deep learning models (particularly CNNs) to leverage spatial patterns and complex features often obscured in 1D signals. While these transformations increase computational demand, they can provide a richer feature space for identifying the electrophysiological dynamics of the heart. The \textit{ECG-biometrics-bench} framework includes a comprehensive \texttt{Representation} module that supports these conversions.

\begin{itemize}
    \item Time-Frequency Domain: The framework supports Spectrograms (e.g., via Short-Time Fourier Transform (STFT)) and Scalograms (e.g., via Continuous Wavelet Transform (CWT)). Spectrograms offer consistent frequency resolution and computational efficiency, whereas scalograms provide a multi-resolution analysis superior for transient feature extraction. The module also includes Mel-Spectrograms, which focus resolution on lower frequency bands where the majority of ECG energy resides. In the context of biometric recognition, both spectrograms and scalograms have proved effective in extracting distinctive features from ECG signals. However, scalograms are more commonly favored over spectrograms. For example, \cite{el2022wavelet, al2024person} used the absolute value of the CWT coefficients of the ECG signal, while \cite{labati2023multicardionet, ammour2023deep} used STFT to create a 2D representation of ECG.
    
    \item Computer Vision Encodings: To capture long-range temporal dependencies and signal dynamics, the framework implements Gramian Angular Fields (GAF) and Recurrence Plots (RP). GAF encodes time-series correlations into polar coordinates, while RP visualizes the recurrence of states in phase space, highlighting patterns particularly useful for noisy bio-signals.
    
    \item Advanced Representations: For specialized analysis, the module provides wrappers for the S-Transform (Stockwell Transform) and the Wigner-Ville Distribution, offering high-resolution time-frequency structures.
\end{itemize}

The \texttt{Representation} module is designed to be stateless and compatible with the framework's segmentation approaches, enabling researchers to conduct ablation studies comparing 1D raw inputs against various 2D image-based encodings to determine the optimal balance between accuracy and system latency.

\subsubsection{Preprocessing Module Control and Orchestration}
The \textit{ECG-biometrics-bench} framework centralizes all signal conditioning within a master \texttt{Preprocessing} class, which orchestrates the transition from raw data to model-ready tensors. This module is designed to be highly configurable, allowing users to control the pipeline through a single interface. The pipeline follows a deterministic four-stage flow.

\begin{enumerate}
    \item Filtering Control: Users can specify the \texttt{filter\_method} (e.g., ``butter'', ``fir'', ``notch'', or ``savgol'') and pass method-specific parameters via \texttt{filter\_kwargs}.
    
    \item Segmentation Selection: The \texttt{mode} parameter toggles between ``beat'' (fiducial) and ``blind'' (non-fiducial) regimes. In beat mode, the module integrates \texttt{NeuroKit2} for R-peak detection and includes an \texttt{align\_peak} logic to center heartbeats precisely, correcting for detection jitter. In blind mode, users control the temporal context and data augmentation through \texttt{window\_s} and \texttt{stride\_s} parameters.
    
    \item Normalization and Resampling: Each segment is individually scaled using either z-score or min-max normalization to mitigate amplitude variations caused by electrode impedance. To ensure compatibility with neural network input layers, the \texttt{resample\_signal} method utilizes Fourier-based resampling to enforce consistent vector lengths.
  
\end{enumerate}

By providing this unified interface, the framework ensures that any specific method change, whether a different filter order or a new window stride, can be tested in a controlled, reproducible manner, making it an ideal tool for comprehensive ablation studies.

\subsection{Data Augmentation}
\label{subsec:augmentation}
Data augmentation aims to increase the amount of available data. Data augmentation also enhances model robustness against noise and variations in different conditions. This is particularly crucial in deep learning contexts, where models require large amounts of training data to generalize well. Unlike traditional machine learning methods, deep learning models internally extract features, eliminating the need for manual feature engineering but increasing data requirements. Data augmentation techniques are widely adopted in fields like computer vision and natural language processing, where they are often an essential part of the training pipeline. While ECG signal processing has seen applications of data augmentation (e.g., in emotion recognition), its use in ECG biometrics remains relatively limited.

Several factors may contribute to the limited adoption of data augmentation in ECG biometrics. Unlike general ECG classification tasks (e.g., detecting arrhythmias or stress levels), biometric applications rely on extracting stable and unique identity-specific features for accurate recognition. In ECG biometrics, augmentation methods that introduce variability can risk distorting the biometric identity if not applied carefully. Moreover, intra-class variability in ECG signals (caused by factors like age, stress, or physical activity) already poses a challenge. Augmenting the data with synthetic variability could further complicate the task, making it harder for models to learn stable identity patterns.

Currently, the application of data augmentation in ECG biometrics is limited; however, there is a potential for future growth and advancement in this area. The very limited available studies employ approaches such as considering signals from different ECG leads (e.g., X, Y, and Z) as separate data to triple training samples \cite{labati2019deep}. Other researchers have utilized multiple random segments from each ECG record \cite{wang2024ecg} or applied direct amplitude scaling to specific wave components to increase training set diversity \cite{hammad2019parallel}. 

To explore the potential of augmentation in ECG biometrics while preserving identity-related characteristics, the framework implements a controlled augmentation pipeline. The implemented transformations aim to simulate realistic acquisition variability while minimizing distortion of subject-specific morphological features. The \texttt{augmentation} module implements several augmentation strategies designed to simulate realistic acquisition variability. All augmentation operations can be applied only during the training phase and are disabled during evaluation to ensure that performance metrics reflect the model's ability to generalize to real signals rather than synthetic perturbations. This design allows the framework to investigate the impact of controlled variability on the robustness of ECG biometric systems while preserving the integrity of identity-related physiological patterns.

\subsection{Model Architectures}
\label{subsec:models}
The \textit{ECG-biometrics-bench} framework integrates several neural architectures to enable systematic benchmarking across different biometric evaluation regimes. The implemented architectures span multiple model families, including convolutional, recurrent, and hybrid. All models are implemented with a flexible output configuration that allows them to operate either as standard classifiers or as feature extractors.

\begin{itemize}

    \item 1D Convolutional Networks: Lightweight convolutional models are widely used in ECG biometrics due to their ability to capture local morphological patterns of the cardiac waveform. The framework implements a standard lightweight 1D-CNN architecture. Additionally, the DeepECG architecture proposed by Labati et al.~\cite{labati2019deep} is included as a representative convolutional model specifically designed for ECG biometric recognition. To explore deeper convolutional representations, a 1D adaptation of the ResNet-18 architecture is also implemented.

\item Recurrent and Hybrid Architectures: To capture temporal dependencies across ECG sequences, the framework includes a bi-directional LSTM model combined with a convolutional front-end for temporal downsampling. In addition, a hybrid CNN-LSTM architecture is implemented, where convolutional layers extract morphological features while recurrent layers model sequential dependencies.

\item 2D Convolutional Models: In addition to time-domain representations, the framework supports image-based ECG representations such as spectrograms and scalograms. For this setting, a 2D ResNet-18 architecture is implemented to process time–frequency representations.

\end{itemize}

\subsection{Evaluation Protocols and Metrics}
\label{subsec:evaluation}

The \textit{ECG-biometrics-bench} framework centralizes all benchmarking logic within the \texttt{run} module, which decouples biometric evaluation protocols from the underlying model architecture. This design allows identical experimental conditions to be applied across different neural models, ensuring reproducibility and fair comparison across architectures.

To systematically evaluate biometric robustness, the framework defines a hierarchy of experimental protocols that progressively introduce two major sources of difficulty: subject generalization (whether identities appear during training) and temporal variability (whether enrollment and probe samples originate from different recording sessions). These regimes can be interpreted as combinations of two orthogonal factors: identity exposure (seen vs. unseen subjects during training) and temporal separation (same-session vs. cross-session evaluation).

\begin{table}[h]
\centering
\caption{Evaluation regimes implemented in the framework.}
\begin{tabular}{lcc}
\hline
\textbf{Regime} & \textbf{Subject Exposure} & \textbf{Temporal Separation} \\
\hline
Closed-set identification & Seen & Same-session \\
Closed-set verification & Seen & Same-session \\
Subject-disjoint identification & Unseen & Same-session \\
Subject-disjoint verification & Unseen & Same-session \\
Cross-session identification & Seen & Cross-session \\
Cross-session verification & Seen & Cross-session \\
Subject-disjoint cross-session identification & Unseen & Cross-session \\
Subject-disjoint cross-session verification & Unseen & Cross-session \\
\hline
\end{tabular}
\end{table}

The evaluation pipeline supports both identification and verification tasks under multiple experimental regimes designed to assess different aspects of biometric robustness. Specifically, the framework implements eight evaluation configurations:

\begin{enumerate}
    \item Closed-set identification: the model is trained and tested on the same identity pool, and the task is to assign each probe ECG to one of the enrolled subjects.

    \item Closed-set verification: enrollment and probe samples originate from identities seen during training, and the system verifies whether two ECG segments belong to the same individual.

    \item Subject-disjoint identification: the model is trained on one set of identities and evaluated on a completely different set, testing whether the learned representation generalizes to unseen subjects.

    \item Subject-disjoint verification: verification is performed on identities not present during training, evaluating the universality of the learned embedding space.

    \item Cross-session identification: enrollment and probe samples originate from different recording sessions of the same individual sets, introducing temporal separation and physiological variability.

    \item Cross-session verification: verification is performed across different recording sessions, evaluating the stability of the biometric system over time.

    \item Subject-disjoint cross-session identification: the model is trained on one identity set and evaluated on unseen identities across different recording sessions, combining subject generalization with temporal variability.

    \item Subject-disjoint cross-session verification: verification is performed on unseen identities across different sessions, testing both subject generalization and long-term biometric stability.
\end{enumerate}

These protocols progressively increase the difficulty of the biometric task. Closed-set scenarios assume that the identities present during testing were already observed during training, whereas subject-disjoint (or open-set) protocols evaluate the ability of the learned representation to generalize to previously unseen individuals. Cross-session configurations further introduce temporal separation between enrollment and probe samples, allowing the framework to assess robustness to physiological changes and template aging.

During training, models are optimized using standard supervised learning procedures. The framework constructs data loaders, manages training loops, and performs validation using configurable hyperparameters. For identification tasks, the model is trained as a multi-class classifier. For verification scenarios, the network acts as a feature extractor that produces compact embedding representations for each ECG segment.

After training, the classification head can be discarded, and the network is used to extract fixed-length embeddings from the penultimate feature layer. These embeddings are then aggregated to construct subject templates for enrollment. For verification experiments, the framework automatically generates genuine and impostor pairs and computes similarity scores between enrollment and probe embeddings. To simulate real-world deployment conditions, threshold-based decision rules can optionally be applied to evaluate overarching verification performance. For identification experiments, similarity scores between each probe embedding and the entire gallery of enrolled templates are organized into a comprehensive score matrix to facilitate Rank-$k$ performance evaluation.

To provide a comprehensive evaluation of biometric performance, the framework reports multiple complementary metrics. For identification tasks, Rank-1 and Rank-5 accuracy are computed. For verification tasks, the framework reports Equal Error Rate (EER), Area Under the ROC Curve (AUC), the decidability index ($d'$), and the True Acceptance Rate at 0.1\% False Acceptance Rate (TAR@0.1\%FAR). These metrics collectively characterize both recognition accuracy and security under strict operating conditions. The experimental evaluation of these protocols and architectures across the integrated ECG datasets is presented in Section~\ref{sec:results}.

\section{Results and Discussion}
\label{sec:results}

We conducted a large-scale benchmark across seven publicly available ECG datasets using the DeepECG architecture \cite{labati2019deep} as a unified baseline model. The objective of this evaluation is not to propose a new recognition model but to systematically analyze how evaluation protocol choice influences the reported performance of ECG biometric systems. Evaluation results are summarized in three tables according to the common characteristics of the datasets. Results for ECG-ID, PTB, and PTB-XL are reported in Table~\ref{tab:comprehensive_results}. Table~\ref{tab:holter_results} presents results for the continuous recording datasets MIT-BIH and NSRDB, while Table~\ref{tab:intervention_results} reports the results for the Heartprint and CYBHi datasets. The benchmark includes multiple evaluation regimes designed to progressively approximate realistic biometric deployment conditions. It should be noted that identification metrics for PTB-XL are omitted from this specific benchmark due to the prohibitive memory constraints associated with generating dense pairwise score matrices for its massive subject cohort.

All models were trained for 250 epochs using the Adam optimizer with a learning rate of 0.001 and a batch size of 256. A fixed training schedule was adopted across all experiments without using a validation split for early stopping. This design choice was made to ensure strict comparability across evaluation protocols and architectures, preventing protocol-dependent bias in model selection. While the proposed framework supports validation-based early stopping and model checkpointing, we intentionally use the final epoch weights to maintain consistency and avoid introducing additional sources of variance across experimental settings. Signal preprocessing followed the standardized pipeline implemented in the proposed framework. This pipeline includes beat-based segmentation (0.2 s before and 0.4 s after the R-peak), band-pass filtering (0.5–40 Hz) using a third-order Butterworth filter, Z-score normalization, and R-peak detection using the Pan–Tompkins algorithm.

To ensure statistical reliability, all experiments were repeated five times with different random seeds. The reported results are expressed as mean $\pm$ standard deviation, where each value represents the average and standard deviation over the five runs. Unless otherwise stated, the results reported in the tables were obtained using cosine similarity for matching, balanced sampling for verification pair generation, heartbeat-level probe fusion with a size of 3 (commonly adopted in the literature), and template fusion using the mean operator. Comprehensive results for all other parameter configurations supported by the framework are available in the accompanying GitHub repository due to space limitations.

\begin{table}[htbp]
\centering
\caption{Comprehensive Evaluation Metrics across ECG-ID, PTB, and PTB-XL Datasets}
\label{tab:comprehensive_results}
\resizebox{\textwidth}{!}{%
\begin{tabular}{l|l|l|cccccc}
\hline
\textbf{Dataset} & \textbf{Data Split} & \textbf{Setting} & \textbf{Rank-1} & \textbf{Rank-5} & \textbf{EER} & \textbf{AUC} & \textbf{D-prime} & \textbf{TAR@0.1\%FAR} \\
\hline

\multirow{14}{*}{\textbf{ECG-ID}} 
 & \multirow{2}{*}{All available} & Closed set & 0.9904 $\pm$ 0.0052 & 0.9977 $\pm$ 0.0016 & 0.0565 $\pm$ 0.0098 & 0.9791 $\pm$ 0.0033 & 3.2966 $\pm$ 0.1324 & 0.7832 $\pm$ 0.0292 \\ \cline{3-9}
 & & Open set & 0.6600 $\pm$ 0.1244 & 0.8478 $\pm$ 0.1282 & 0.1736 $\pm$ 0.0195 & 0.9027 $\pm$ 0.0170 & 1.9690 $\pm$ 0.1869 & 0.3487 $\pm$ 0.1491 \\ \cline{2-9}
 & \multirow{2}{*}{Single session} & Closed set & 0.9664 $\pm$ 0.0088 & 0.9984 $\pm$ 0.0036 & 0.0659 $\pm$ 0.0077 & 0.9809 $\pm$ 0.0034 & 2.8371 $\pm$ 0.1057 & 0.6439 $\pm$ 0.0794 \\ \cline{3-9}
 & & Open set & 0.6078 $\pm$ 0.0490 & 0.8782 $\pm$ 0.0570 & 0.1944 $\pm$ 0.0263 & 0.8871 $\pm$ 0.0304 & 1.7828 $\pm$ 0.2660 & 0.2718 $\pm$ 0.0656 \\ \cline{2-9}
 & \multirow{2}{*}{Single cross session} & Closed set & 0.8448 $\pm$ 0.0121 & 0.9433 $\pm$ 0.0048 & 0.0870 $\pm$ 0.0014 & 0.9714 $\pm$ 0.0012 & 2.7410 $\pm$ 0.0482 & 0.5694 $\pm$ 0.0624 \\ \cline{3-9}
 & & Open set & 0.9275 $\pm$ 0.0251 & 0.9762 $\pm$ 0.0248 & 0.0894 $\pm$ 0.0180 & 0.9722 $\pm$ 0.0087 & 2.6129 $\pm$ 0.1738 & 0.5077 $\pm$ 0.0643 \\ \cline{2-9}
 & \multirow{2}{*}{SS Short-term} & Closed set & 0.7753 $\pm$ 0.0190 & 0.9161 $\pm$ 0.0067 & 0.0861 $\pm$ 0.0018 & 0.9712 $\pm$ 0.0007 & 2.7201 $\pm$ 0.0279 & 0.5184 $\pm$ 0.0624 \\ \cline{3-9}
 & & Open set & 0.8955 $\pm$ 0.0410 & 0.9797 $\pm$ 0.0279 & 0.0854 $\pm$ 0.0141 & 0.9715 $\pm$ 0.0087 & 2.6227 $\pm$ 0.1226 & 0.4814 $\pm$ 0.1131 \\ \cline{2-9}
 & \multirow{2}{*}{LLO Short-term} & Closed set & 0.9175 $\pm$ 0.0085 & 0.9805 $\pm$ 0.0017 & 0.0573 $\pm$ 0.0045 & 0.9824 $\pm$ 0.0011 & 2.9922 $\pm$ 0.0711 & 0.5992 $\pm$ 0.0579 \\ \cline{3-9}
 & & Open set & 0.9443 $\pm$ 0.0374 & 0.9963 $\pm$ 0.0053 & 0.0728 $\pm$ 0.0274 & 0.9764 $\pm$ 0.0123 & 2.7668 $\pm$ 0.2691 & 0.5578 $\pm$ 0.1763 \\ \cline{2-9}
 & \multirow{2}{*}{SS Long-term} & Closed set & 0.8214 $\pm$ 0.0187 & 0.9769 $\pm$ 0.0065 & 0.0858 $\pm$ 0.0075 & 0.9665 $\pm$ 0.0061 & 2.3568 $\pm$ 0.1297 & 0.5698 $\pm$ 0.1053 \\ \cline{3-9}
 & & Open set & 0.9143 $\pm$ 0.0757 & 1.0000 $\pm$ 0.0000 & 0.1744 $\pm$ 0.0968 & 0.9048 $\pm$ 0.0872 & 1.8867 $\pm$ 0.8240 & 0.3363 $\pm$ 0.1908 \\ \cline{2-9}
 & \multirow{2}{*}{LLO Long-term} & Closed set & 0.8893 $\pm$ 0.0109 & 0.9678 $\pm$ 0.0097 & 0.0797 $\pm$ 0.0069 & 0.9696 $\pm$ 0.0038 & 2.5635 $\pm$ 0.1120 & 0.7044 $\pm$ 0.0936 \\ \cline{3-9}
 & & Open set & 0.8978 $\pm$ 0.0628 & 1.0000 $\pm$ 0.0000 & 0.1613 $\pm$ 0.0824 & 0.9217 $\pm$ 0.0481 & 1.9578 $\pm$ 0.5664 & 0.3149 $\pm$ 0.1193 \\
\hline
\hline

\multirow{14}{*}{\textbf{PTB}} 
 & \multirow{2}{*}{All available} & Closed set & 0.9968 $\pm$ 0.0008 & 0.9994 $\pm$ 0.0003 & 0.0207 $\pm$ 0.0026 & 0.9949 $\pm$ 0.0012 & 4.8041 $\pm$ 0.1761 & 0.9511 $\pm$ 0.0057 \\ \cline{3-9}
 & & Open set & 0.7041 $\pm$ 0.0247 & 0.8675 $\pm$ 0.0081 & 0.1459 $\pm$ 0.0082 & 0.9033 $\pm$ 0.0051 & 2.0659 $\pm$ 0.0511 & 0.5290 $\pm$ 0.0262 \\ \cline{2-9}
 & \multirow{2}{*}{Single session} & Closed set & 0.9972 $\pm$ 0.0016 & 0.9995 $\pm$ 0.0002 & 0.0192 $\pm$ 0.0020 & 0.9950 $\pm$ 0.0009 & 4.6046 $\pm$ 0.1271 & 0.9520 $\pm$ 0.0067 \\ \cline{3-9}
 & & Open set & 0.7866 $\pm$ 0.0328 & 0.8650 $\pm$ 0.0210 & 0.1475 $\pm$ 0.0260 & 0.9007 $\pm$ 0.0156 & 1.9921 $\pm$ 0.1229 & 0.5362 $\pm$ 0.0913 \\ \cline{2-9}
 & \multirow{2}{*}{Single cross session} & Closed set & 0.6734 $\pm$ 0.0368 & 0.8646 $\pm$ 0.0244 & 0.0910 $\pm$ 0.0037 & 0.9652 $\pm$ 0.0031 & 2.7288 $\pm$ 0.0780 & 0.4747 $\pm$ 0.0776 \\ \cline{3-9}
 & & Open set & 0.7330 $\pm$ 0.1023 & 0.9611 $\pm$ 0.0081 & 0.1203 $\pm$ 0.0281 & 0.9501 $\pm$ 0.0203 & 2.4377 $\pm$ 0.3193 & 0.3656 $\pm$ 0.1207 \\ \cline{2-9}
 & \multirow{2}{*}{SS Short-term} & Closed set & 0.9147 $\pm$ 0.0014 & 0.9180 $\pm$ 0.0007 & 0.1011 $\pm$ 0.0042 & 0.9564 $\pm$ 0.0110 & 2.5296 $\pm$ 0.1193 & 0.7394 $\pm$ 0.1230 \\ \cline{3-9}
 & & Open set & 0.8210 $\pm$ 0.1255 & 1.0000 $\pm$ 0.0000 & 0.1936 $\pm$ 0.0921 & 0.8909 $\pm$ 0.0728 & 1.8177 $\pm$ 0.7796 & 0.7081 $\pm$ 0.1189 \\ \cline{2-9}
 & \multirow{2}{*}{LLO Short-term} & Closed set & 0.8959 $\pm$ 0.0017 & 0.9063 $\pm$ 0.0062 & 0.0984 $\pm$ 0.0043 & 0.9688 $\pm$ 0.0108 & 2.7168 $\pm$ 0.1841 & 0.7736 $\pm$ 0.0440 \\ \cline{3-9}
 & & Open set & 0.8048 $\pm$ 0.1124 & 1.0000 $\pm$ 0.0000 & 0.1898 $\pm$ 0.0906 & 0.8922 $\pm$ 0.0596 & 1.7612 $\pm$ 0.5726 & 0.7455 $\pm$ 0.1104 \\ \cline{2-9}
 & \multirow{2}{*}{SS Long-term} & Closed set & 0.4982 $\pm$ 0.0214 & 0.7303 $\pm$ 0.0212 & 0.1326 $\pm$ 0.0076 & 0.9365 $\pm$ 0.0053 & 2.1866 $\pm$ 0.0640 & 0.2855 $\pm$ 0.0226 \\ \cline{3-9}
 & & Open set & 0.6524 $\pm$ 0.0482 & 0.8921 $\pm$ 0.0262 & 0.1549 $\pm$ 0.0158 & 0.9188 $\pm$ 0.0139 & 1.9934 $\pm$ 0.1312 & 0.2478 $\pm$ 0.0576 \\ \cline{2-9}
 & \multirow{2}{*}{LLO Long-term} & Closed set & 0.7986 $\pm$ 0.0279 & 0.9382 $\pm$ 0.0138 & 0.0709 $\pm$ 0.0104 & 0.9763 $\pm$ 0.0043 & 3.0372 $\pm$ 0.1189 & 0.6108 $\pm$ 0.0677 \\ \cline{3-9}
 & & Open set & 0.8083 $\pm$ 0.0748 & 0.9633 $\pm$ 0.0289 & 0.1050 $\pm$ 0.0290 & 0.9521 $\pm$ 0.0160 & 2.4846 $\pm$ 0.2285 & 0.3619 $\pm$ 0.0935 \\
\hline
\hline

\multirow{14}{*}{\textbf{PTB-XL}} 
 & \multirow{2}{*}{All available} & Closed set & - - - & - - - & 0.0464 $\pm$ 0.0050 & 0.9770 $\pm$ 0.0033 & 3.7101 $\pm$ 0.1716 & 0.8596 $\pm$ 0.0258 \\ \cline{3-9}
 & & Open set & - - - & - - - & 0.2254 $\pm$ 0.0108 & 0.8401 $\pm$ 0.0097 & 1.6454 $\pm$ 0.0740 & 0.4512 $\pm$ 0.0360 \\ \cline{2-9}
 & \multirow{2}{*}{Single session} & Closed set & - - - & - - - & 0.0440 $\pm$ 0.0048 & 0.9785 $\pm$ 0.0040 & 3.6207 $\pm$ 0.1678 & 0.8846 $\pm$ 0.0289 \\ \cline{3-9}
 & & Open set & - - - & - - - & 0.2364 $\pm$ 0.0150 & 0.8360 $\pm$ 0.0087 & 1.6012 $\pm$ 0.0628 & 0.4402 $\pm$ 0.0342 \\ \cline{2-9}
 & \multirow{2}{*}{Single cross session} & Closed set & - - - & - - - & 0.1397 $\pm$ 0.0037 & 0.9297 $\pm$ 0.0042 & 2.2139 $\pm$ 0.0633 & 0.2816 $\pm$ 0.0437 \\ \cline{3-9}
 & & Open set & - - - & - - - & 0.1436 $\pm$ 0.0069 & 0.9262 $\pm$ 0.0057 & 2.1634 $\pm$ 0.0821 & 0.2561 $\pm$ 0.0657 \\ \cline{2-9}
 & \multirow{2}{*}{SS Short-term} & Closed set & - - - & - - - & 0.1379 $\pm$ 0.0162 & 0.9270 $\pm$ 0.0074 & 1.8766 $\pm$ 0.0680 & 0.3452 $\pm$ 0.0653 \\ \cline{3-9}
 & & Open set & - - - & - - - & 0.2428 $\pm$ 0.0694 & 0.8594 $\pm$ 0.0491 & 1.3739 $\pm$ 0.2910 & 0.1780 $\pm$ 0.0437 \\ \cline{2-9}
 & \multirow{2}{*}{LLO Short-term} & Closed set & - - - & - - - & 0.1158 $\pm$ 0.0162 & 0.9481 $\pm$ 0.0089 & 1.9747 $\pm$ 0.1166 & 0.4552 $\pm$ 0.0563 \\ \cline{3-9}
 & & Open set & - - - & - - - & 0.1673 $\pm$ 0.0383 & 0.9107 $\pm$ 0.0350 & 1.6877 $\pm$ 0.2408 & 0.3576 $\pm$ 0.1094 \\ \cline{2-9}
 & \multirow{2}{*}{SS Long-term} & Closed set & - - - & - - - & 0.1383 $\pm$ 0.0037 & 0.9300 $\pm$ 0.0021 & 2.2299 $\pm$ 0.0485 & 0.2611 $\pm$ 0.0170 \\ \cline{3-9}
 & & Open set & - - - & - - - & 0.1367 $\pm$ 0.0113 & 0.9313 $\pm$ 0.0085 & 2.2246 $\pm$ 0.0949 & 0.2726 $\pm$ 0.0337 \\ \cline{2-9}
 & \multirow{2}{*}{LLO Long-term} & Closed set & - - - & - - - & 0.1203 $\pm$ 0.0060 & 0.9413 $\pm$ 0.0053 & 2.3954 $\pm$ 0.0888 & 0.3150 $\pm$ 0.0609 \\ \cline{3-9}
 & & Open set & - - - & - - - & 0.1220 $\pm$ 0.0132 & 0.9417 $\pm$ 0.0089 & 2.4193 $\pm$ 0.1616 & 0.2947 $\pm$ 0.0625 \\
\hline

\end{tabular}%
}

\vspace{0.15cm} 
\raggedright \footnotesize 
\textit{Note:} \textbf{SS}: Single-Shot; \textbf{LLO}: Leave-Last-Out. In this evaluation framework, \textbf{Open set} refers strictly to the subject-disjoint protocol, whereas \textbf{Closed set} evaluates subjects already seen during enrollment.

\end{table}

\begin{table}[htbp]
\centering
\caption{Comprehensive Evaluation Metrics across MIT-BIH and NSRDB Datasets}
\label{tab:holter_results}
\resizebox{\textwidth}{!}{%
\begin{tabular}{l|l|l|cccccc}
\hline
\textbf{Dataset} & \textbf{Data Split} & \textbf{Setting} & \textbf{Rank-1} & \textbf{Rank-5} & \textbf{EER} & \textbf{AUC} & \textbf{D-prime} & \textbf{TAR@FAR} \\
\hline

\multirow{8}{*}{\textbf{MIT-BIH}} 
 & \multirow{2}{*}{Single segment} & Closed set & 0.5338 $\pm$ 0.0184 & 1.0000 $\pm$ 0.0000 & 0.0236 $\pm$ 0.0021 & 0.9916 $\pm$ 0.0012 & 5.5953 $\pm$ 0.2413 & 0.0846 $\pm$ 0.0369 \\ \cline{3-9}
 & & Open set & 0.8428 $\pm$ 0.0496 & 0.9664 $\pm$ 0.0286 & 0.1482 $\pm$ 0.0337 & 0.9202 $\pm$ 0.0250 & 2.3434 $\pm$ 0.3472 & 0.5424 $\pm$ 0.1655 \\ \cline{2-9}
 & \multirow{2}{*}{Short-term} & Closed set & 0.6695 $\pm$ 0.0042 & 0.9884 $\pm$ 0.0055 & 0.0329 $\pm$ 0.0051 & 0.9874 $\pm$ 0.0017 & 5.0653 $\pm$ 0.2051 & 0.1231 $\pm$ 0.0342 \\ \cline{3-9}
 & & Open set & 0.9540 $\pm$ 0.0476 & 0.9981 $\pm$ 0.0019 & 0.0522 $\pm$ 0.0204 & 0.9852 $\pm$ 0.0062 & 4.1317 $\pm$ 0.5104 & 0.6556 $\pm$ 0.2602 \\ \cline{2-9}
 & \multirow{2}{*}{Long-term} & Closed set & 0.9014 $\pm$ 0.0053 & 0.9655 $\pm$ 0.0044 & 0.0822 $\pm$ 0.0068 & 0.9744 $\pm$ 0.0020 & 3.3532 $\pm$ 0.0830 & 0.7160 $\pm$ 0.0111 \\ \cline{3-9}
 & & Open set & 0.9158 $\pm$ 0.0441 & 0.9914 $\pm$ 0.0154 & 0.1239 $\pm$ 0.0345 & 0.9489 $\pm$ 0.0175 & 2.7253 $\pm$ 0.4063 & 0.6456 $\pm$ 0.1564 \\ \cline{2-9}
 & \multirow{2}{*}{Multi-shot} & Closed set & 0.9031 $\pm$ 0.0080 & 0.9681 $\pm$ 0.0017 & 0.0833 $\pm$ 0.0057 & 0.9742 $\pm$ 0.0033 & 3.3440 $\pm$ 0.0853 & 0.7070 $\pm$ 0.0478 \\ \cline{3-9}
 & & Open set & 0.9173 $\pm$ 0.0455 & 0.9905 $\pm$ 0.0154 & 0.1343 $\pm$ 0.0335 & 0.9456 $\pm$ 0.0153 & 2.6384 $\pm$ 0.2933 & 0.6351 $\pm$ 0.1514 \\
\hline
\hline

\multirow{8}{*}{\textbf{NSRDB}} 
 & \multirow{2}{*}{Single segment} & Closed set & 1.0000 $\pm$ 0.0000 & 1.0000 $\pm$ 0.0000 & 0.0056 $\pm$ 0.0021 & 0.9993 $\pm$ 0.0002 & 3.8083 $\pm$ 0.2098 & 0.9894 $\pm$ 0.0038 \\ \cline{3-9}
 & & Open set & 0.8158 $\pm$ 0.1785 & 1.0000 $\pm$ 0.0000 & 0.1846 $\pm$ 0.0782 & 0.8883 $\pm$ 0.1044 & 1.9584 $\pm$ 0.7741 & 0.3534 $\pm$ 0.1017 \\ \cline{2-9}
 & \multirow{2}{*}{Short-term} & Closed set & 0.9964 $\pm$ 0.0005 & 1.0000 $\pm$ 0.0000 & 0.0077 $\pm$ 0.0017 & 0.9989 $\pm$ 0.0004 & 4.6226 $\pm$ 0.1443 & 0.9730 $\pm$ 0.0075 \\ \cline{3-9}
 & & Open set & 0.9910 $\pm$ 0.0119 & 1.0000 $\pm$ 0.0000 & 0.0532 $\pm$ 0.0393 & 0.9868 $\pm$ 0.0103 & 3.3815 $\pm$ 1.0741 & 0.6862 $\pm$ 0.2822 \\ \cline{2-9}
 & \multirow{2}{*}{Long-term} & Closed set & 0.0894 $\pm$ 0.0029 & 0.4418 $\pm$ 0.0055 & 0.4950 $\pm$ 0.0102 & 0.5038 $\pm$ 0.0135 & 0.0295 $\pm$ 0.0388 & 0.0015 $\pm$ 0.0013 \\ \cline{3-9}
 & & Open set & 0.3390 $\pm$ 0.0492 & 1.0000 $\pm$ 0.0000 & 0.4944 $\pm$ 0.0227 & 0.5088 $\pm$ 0.0211 & 0.0388 $\pm$ 0.0414 & 0.0035 $\pm$ 0.0052 \\ \cline{2-9}
 & \multirow{2}{*}{Multi-shot} & Closed set & 0.0971 $\pm$ 0.0227 & 0.4402 $\pm$ 0.0218 & 0.4929 $\pm$ 0.0167 & 0.5113 $\pm$ 0.0183 & 0.0678 $\pm$ 0.0291 & 0.0032 $\pm$ 0.0035 \\ \cline{3-9}
 & & Open set & 0.3421 $\pm$ 0.0596 & 1.0000 $\pm$ 0.0000 & 0.4862 $\pm$ 0.0225 & 0.5160 $\pm$ 0.0294 & 0.0733 $\pm$ 0.0759 & 0.0010 $\pm$ 0.0013 \\
\hline
\end{tabular}%
}

\vspace{0.15cm}
\raggedright \footnotesize 
\textit{Note:} In this evaluation framework, \textbf{Open set} refers strictly to the subject-disjoint protocol, whereas \textbf{Closed set} evaluates subjects already seen during enrollment.

\end{table}

\begin{table}[htbp]
\centering
\caption{Comprehensive Evaluation Metrics across CYBHi and Heartprint Datasets}
\label{tab:intervention_results}
\resizebox{\textwidth}{!}{%
\begin{tabular}{l|l|l|cccccc}
\hline
\textbf{Dataset} & \textbf{Data Split} & \textbf{Setting} & \textbf{Rank-1} & \textbf{Rank-5} & \textbf{EER} & \textbf{AUC} & \textbf{D-prime} & \textbf{TAR@FAR} \\
\hline

\multirow{10}{*}{\textbf{CYBHi}} 
 & \multirow{2}{*}{Baseline CI} & Closed set & 0.9621 $\pm$ 0.0034 & 0.9965 $\pm$ 0.0029 & 0.1205 $\pm$ 0.0043 & 0.9360 $\pm$ 0.0032 & 2.3640 $\pm$ 0.0694 & 0.6265 $\pm$ 0.0464 \\ \cline{3-9}
 & & Open set & 0.3945 $\pm$ 0.0448 & 0.6093 $\pm$ 0.1082 & 0.3920 $\pm$ 0.0609 & 0.6707 $\pm$ 0.0653 & 0.7197 $\pm$ 0.2351 & 0.0968 $\pm$ 0.0854 \\ \cline{2-9}
 & \multirow{2}{*}{Baseline S1} & Closed set & 0.9843 $\pm$ 0.0052 & 0.9989 $\pm$ 0.0006 & 0.0875 $\pm$ 0.0039 & 0.9685 $\pm$ 0.0027 & 2.7870 $\pm$ 0.0825 & 0.5773 $\pm$ 0.0532 \\ \cline{3-9}
 & & Open set & 0.4062 $\pm$ 0.0653 & 0.7461 $\pm$ 0.0612 & 0.3188 $\pm$ 0.0256 & 0.7483 $\pm$ 0.0244 & 0.9893 $\pm$ 0.0884 & 0.1178 $\pm$ 0.0290 \\ \cline{2-9}
 & \multirow{2}{*}{Intervention A1} & Closed set & 0.3521 $\pm$ 0.0151 & 0.4602 $\pm$ 0.0105 & 0.4185 $\pm$ 0.0087 & 0.6542 $\pm$ 0.0089 & 0.6773 $\pm$ 0.0527 & 0.2353 $\pm$ 0.0254 \\ \cline{3-9}
 & & Open set & 0.3935 $\pm$ 0.0638 & 0.6619 $\pm$ 0.0404 & 0.4413 $\pm$ 0.0355 & 0.6344 $\pm$ 0.0370 & 0.5911 $\pm$ 0.1210 & 0.2142 $\pm$ 0.0714 \\ \cline{2-9}
 & \multirow{2}{*}{Intervention A2} & Closed set & 0.3008 $\pm$ 0.0198 & 0.4357 $\pm$ 0.0101 & 0.4182 $\pm$ 0.0105 & 0.6494 $\pm$ 0.0068 & 0.6540 $\pm$ 0.0409 & 0.2020 $\pm$ 0.0211 \\ \cline{3-9}
 & & Open set & 0.3596 $\pm$ 0.0589 & 0.6400 $\pm$ 0.0582 & 0.4479 $\pm$ 0.0250 & 0.6272 $\pm$ 0.0261 & 0.5545 $\pm$ 0.1128 & 0.1316 $\pm$ 0.0531 \\ \cline{2-9}
 & \multirow{2}{*}{Long-term} & Closed set & 0.6000 $\pm$ 0.0284 & 0.7994 $\pm$ 0.0217 & 0.1974 $\pm$ 0.0078 & 0.8711 $\pm$ 0.0105 & 1.6418 $\pm$ 0.1038 & 0.2412 $\pm$ 0.0451 \\ \cline{3-9}
 & & Open set & 0.7286 $\pm$ 0.0877 & 0.9435 $\pm$ 0.0375 & 0.1707 $\pm$ 0.0405 & 0.8993 $\pm$ 0.0375 & 1.8503 $\pm$ 0.3367 & 0.2474 $\pm$ 0.0834 \\
\hline
\hline

\multirow{14}{*}{\textbf{Heartprint}} 
 & \multirow{2}{*}{Single session} & Closed set & 0.9833 $\pm$ 0.0067 & 0.9982 $\pm$ 0.0013 & 0.0443 $\pm$ 0.0051 & 0.9860 $\pm$ 0.0029 & 3.6546 $\pm$ 0.1410 & 0.7453 $\pm$ 0.0519 \\ \cline{3-9}
 & & Open set & 0.5902 $\pm$ 0.0376 & 0.7741 $\pm$ 0.0410 & 0.2008 $\pm$ 0.0264 & 0.8740 $\pm$ 0.0299 & 1.8025 $\pm$ 0.2153 & 0.3211 $\pm$ 0.0367 \\ \cline{2-9}
 & \multirow{2}{*}{Short-term} & Closed set & 0.4549 $\pm$ 0.0126 & 0.7254 $\pm$ 0.0074 & 0.1273 $\pm$ 0.0069 & 0.9436 $\pm$ 0.0044 & 2.4507 $\pm$ 0.0687 & 0.2578 $\pm$ 0.0534 \\ \cline{3-9}
 & & Open set & 0.5708 $\pm$ 0.0628 & 0.8828 $\pm$ 0.0342 & 0.1422 $\pm$ 0.0152 & 0.9362 $\pm$ 0.0034 & 2.3250 $\pm$ 0.0711 & 0.2254 $\pm$ 0.0446 \\ \cline{2-9}
 & \multirow{2}{*}{Short-term (Rev.)} & Closed set & 0.4658 $\pm$ 0.0113 & 0.7300 $\pm$ 0.0204 & 0.1260 $\pm$ 0.0092 & 0.9461 $\pm$ 0.0077 & 2.4985 $\pm$ 0.1211 & 0.2192 $\pm$ 0.0965 \\ \cline{3-9}
 & & Open set & 0.5742 $\pm$ 0.0591 & 0.8565 $\pm$ 0.0585 & 0.1474 $\pm$ 0.0220 & 0.9270 $\pm$ 0.0122 & 2.2288 $\pm$ 0.1649 & 0.2030 $\pm$ 0.0391 \\ \cline{2-9}
 & \multirow{2}{*}{Cognitive} & Closed set & 0.4049 $\pm$ 0.0175 & 0.6954 $\pm$ 0.0143 & 0.1688 $\pm$ 0.0024 & 0.9002 $\pm$ 0.0044 & 1.9242 $\pm$ 0.0552 & 0.1691 $\pm$ 0.0285 \\ \cline{3-9}
 & & Open set & 0.5735 $\pm$ 0.0282 & 0.8976 $\pm$ 0.0570 & 0.1744 $\pm$ 0.0262 & 0.8980 $\pm$ 0.0320 & 1.9013 $\pm$ 0.2865 & 0.1651 $\pm$ 0.0674 \\ \cline{2-9}
 & \multirow{2}{*}{Cognitive (S1-S2-S3R)} & Closed set & 0.4076 $\pm$ 0.0060 & 0.7128 $\pm$ 0.0140 & 0.1690 $\pm$ 0.0064 & 0.9002 $\pm$ 0.0069 & 1.9251 $\pm$ 0.0806 & 0.1728 $\pm$ 0.0381 \\ \cline{3-9}
 & & Open set & 0.5748 $\pm$ 0.0536 & 0.8994 $\pm$ 0.0502 & 0.1691 $\pm$ 0.0364 & 0.8985 $\pm$ 0.0386 & 1.9383 $\pm$ 0.3623 & 0.1479 $\pm$ 0.1003 \\ \cline{2-9}
 & \multirow{2}{*}{Long-term} & Closed set & 0.3879 $\pm$ 0.0119 & 0.6952 $\pm$ 0.0168 & 0.1840 $\pm$ 0.0108 & 0.8865 $\pm$ 0.0062 & 1.7877 $\pm$ 0.0546 & 0.1350 $\pm$ 0.0244 \\ \cline{3-9}
 & & Open set & 0.6079 $\pm$ 0.1067 & 0.8922 $\pm$ 0.0743 & 0.1990 $\pm$ 0.0411 & 0.8716 $\pm$ 0.0401 & 1.7123 $\pm$ 0.3629 & 0.0802 $\pm$ 0.0513 \\ \cline{2-9}
 & \multirow{2}{*}{Long-term (S1-S2-S3L)} & Closed set & 0.3910 $\pm$ 0.0116 & 0.6864 $\pm$ 0.0248 & 0.1866 $\pm$ 0.0086 & 0.8871 $\pm$ 0.0068 & 1.7956 $\pm$ 0.0517 & 0.1408 $\pm$ 0.0289 \\ \cline{3-9}
 & & Open set & 0.6018 $\pm$ 0.0965 & 0.8761 $\pm$ 0.0776 & 0.2048 $\pm$ 0.0460 & 0.8709 $\pm$ 0.0465 & 1.7004 $\pm$ 0.4003 & 0.0654 $\pm$ 0.0366 \\
\hline
\end{tabular}%
}

\vspace{0.15cm}
\raggedright \footnotesize 
\textit{Note:} \textbf{CYBHi}: \textit{A1} denotes Physical Exercise Intervention; \textit{A2} denotes Mental Stress Intervention. \textbf{Heartprint}: \textit{Rev.} denotes Reversed order of evaluation. In this framework, \textbf{Open set} refers strictly to the subject-disjoint protocol, whereas \textbf{Closed set} evaluates subjects already seen during enrollment.

\end{table}

\subsection{Random Split Baselines: Artificially Inflated Performance}
\label{subsec:random_split}

The commonly used random intra-session protocol is first evaluated, in which enrollment and probe samples are randomly selected from the same recording session. Although this protocol is widely adopted in the ECG biometric literature, it does not represent realistic deployment conditions. Under this protocol, nearly all datasets demonstrate extremely high performance. For instance, Rank-1 identification reaches 100\% on NSRDB and 99.68\% on PTB, while verification error rates fall below 5\% in several datasets. Even datasets with lower signal-to-noise ratios, such as CYBHi, achieve nearly 97\% Rank-1 accuracy in both baseline sessions.

\begin{figure}
    \centering
    \includegraphics[width=0.75\linewidth]{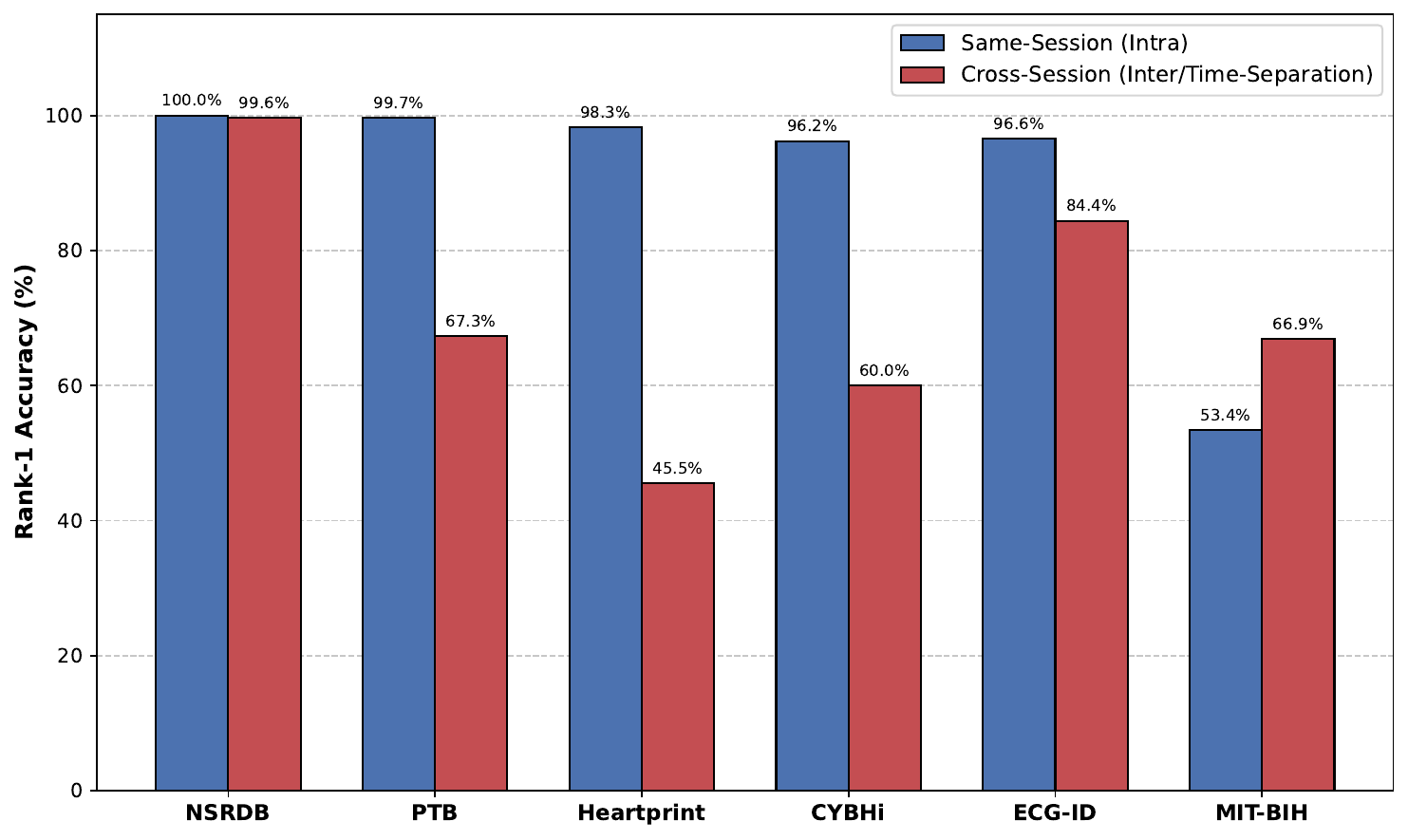}
    \caption{Rank-1 Comparison Between Intra-Session and Inter-Session Evaluation}
    \label{fig:performance_decline_rank1}
\end{figure}

\begin{figure}
    \centering
    \includegraphics[width=0.75\linewidth]{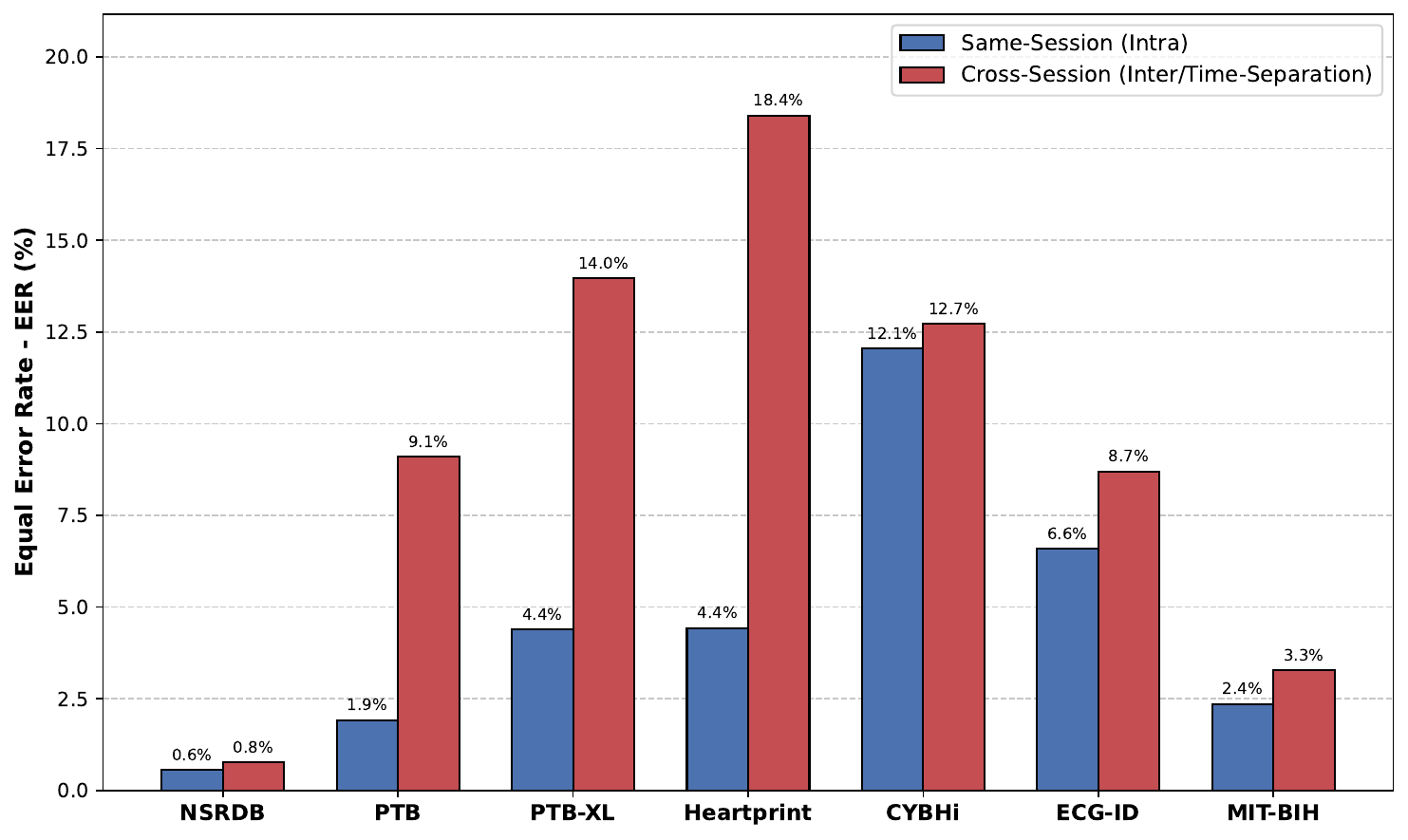}
    \caption{EER Comparison Between Intra-Session and Inter-Session Evaluation}
    \label{fig:performance_degradation_eer}
\end{figure}

However, these results primarily reflect strong temporal correlations within recordings. Adjacent ECG segments exhibit nearly identical morphology and differ due to minor noise variations. Consequently, models tend to memorize session-specific signal characteristics instead of learning robust identity features. This limitation becomes apparent under more realistic evaluation protocols, in which enrollment and probe samples are temporally separated. Across datasets, performance declines substantially when random splitting is replaced by cross-session evaluation, indicating that random splits consistently overestimate the effectiveness of ECG biometrics. For instance, Heartprint achieves nearly 98\% identification accuracy under intra-session random splitting; however, its performance decreases to 45.5\% under long-term evaluation, where sessions are separated by an average interval of 47.5 days. Comparable trends are observed in PTB and ECG-ID, with Rank-1 identification decreasing by more than 30\% and 12\%, respectively, even when the separated sessions are recorded on the same day. These results indicate that random intra-session evaluation does not adequately assess the true biometric capability of ECG signals. Therefore, temporally separated protocols should be considered the minimum standard for reliable evaluation. Figure~\ref{fig:performance_decline_rank1} and Figure~\ref{fig:performance_degradation_eer} illustrate the decline in performance when training and testing data are selected from different sessions across various datasets. These figures present cross-session or short-term scenarios, except for the CYBHi dataset, where an explicit short-term scenario without physical and physiological intervention is unavailable. In this case, the long-term results are shown for comparison with the intra-session scenario.

An anomaly is observed in the MIT-BIH dataset, where Rank-1 identification accuracy improves from 53.4\% in the single-segment baseline to 67.0\% in the short-term split, while the verification EER slightly degrades from 2.4\% to 3.3\%. This contradictory outcome can be primarily due to the continuous nature of the dataset. Unlike datasets with distinct acquisition sessions, MIT-BIH comprises a single uninterrupted recording, so session splits are artificially defined and do not involve electrode repositioning or hardware changes. Consequently, differences between splits mainly reflect temporal variations within the same recording rather than genuine session shifts. Furthermore, variations in signal quality across temporal segments may contribute to this effect, as certain intervals may provide cleaner and more discriminative patterns for identification. A similar trend is observed in the NSRDB dataset, which also consists of continuous recordings; in this case, artificially separating sessions for training and testing has minimal impact on Rank-1 accuracy.

\subsection{Physiological State Robustness}

The CYBHi dataset facilitates a comprehensive analysis of biometric system robustness under physiological changes, such as physical exercise and cognitive stress. These experimental conditions replicate realistic scenarios in which biometric systems must identify users across varying physiological states. The results demonstrate a substantial decline in identification and verification performance when enrollment and probe samples are collected under differing physiological conditions. On the CYBHi dataset, baseline intra-session evaluation achieves approximately 98\% Rank-1 accuracy, whereas cross-session intervention protocols reduce performance to approximately 35\% for exercise and 30\% for mental stress. Verification error rates also increase, exceeding 40\% EER. These results suggest that current ECG biometric systems are highly sensitive to changes in physiological state.

A comparable trend is observed in the Heartprint dataset when comparing the Rank-1 results of the single-session task and the cross-session task, where S1 and S3R serve as the training and testing sessions, respectively. The performance decreases from 98\% to 40\%. Notably, in the Heartprint dataset, this decline is attributable not only to physical changes in the subjects but also to the extended time interval between sessions (1054 days on average).

\subsection{Long-Term Stability and Template Aging}
\label{subsec:template_aging}

\begin{figure}[htbp]
    \centering
    \includegraphics[width=0.75\linewidth]{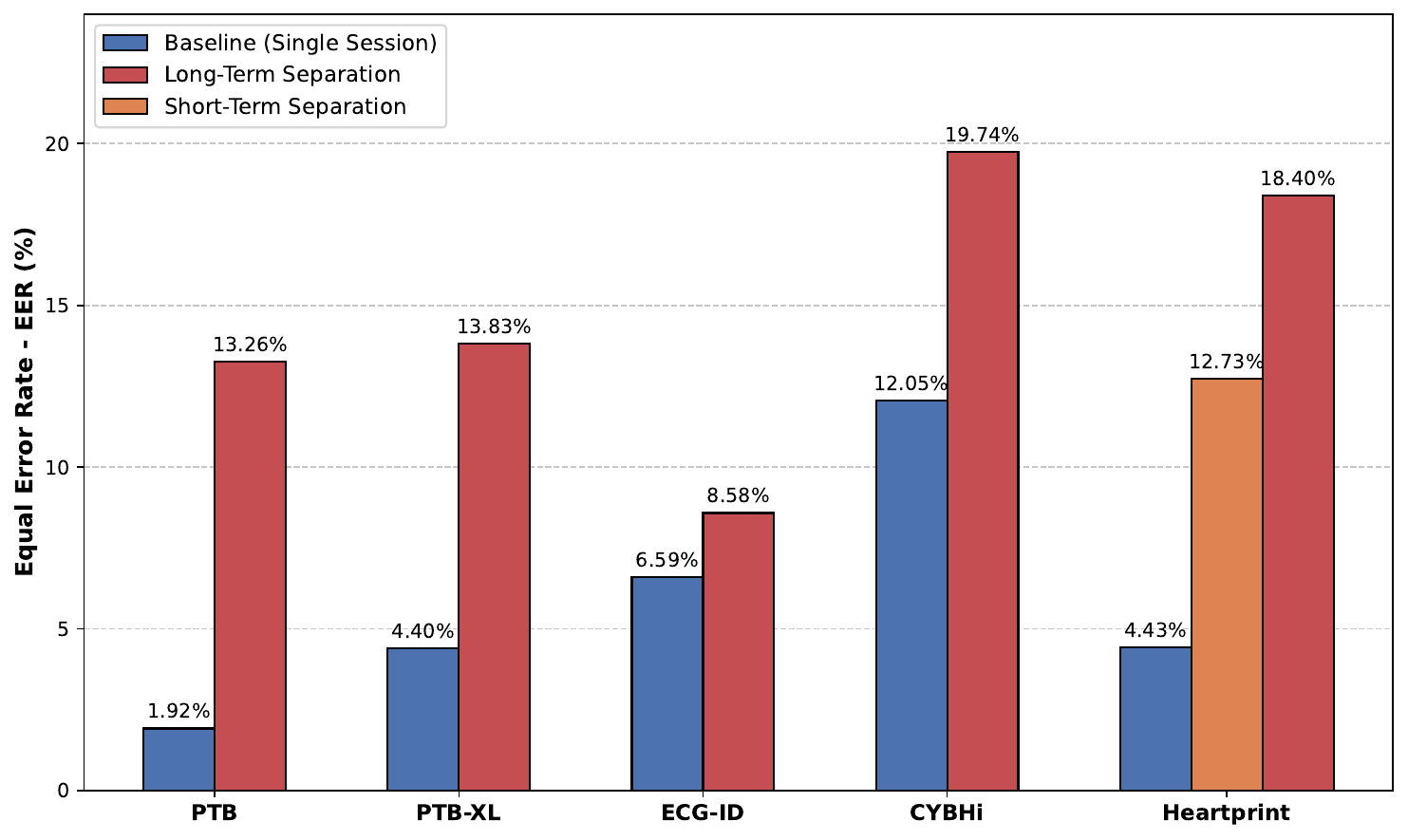}
    \caption{Verification Degradation Over Time: Closed-Set EER Comparison Between Intra-Session Baseline and Long-Term Evaluation.}
    \label{fig:long_term_eer_stability}
\end{figure}

Long-term evaluation protocols impose significant temporal separation between enrollment and probe samples, thereby explicitly revealing the effects of template aging over extended durations. As shown in Figure~\ref{fig:long_term_eer_stability}, extended temporal intervals consistently result in notable degradation of verification performance across various datasets. Although baseline intra-session evaluations produce highly optimistic EER, the introduction of temporal separation exposes the inherent vulnerability of static biometric templates.

On the PTB dataset, the closed-set EER increases from 1.9\% under same-session evaluation to 13.3\% when enrollment and probe recordings are separated by extended time intervals. The Heartprint dataset most clearly demonstrates the progressive nature of template aging, with verification error rising from 4.4\% in the single-session baseline to 12.7\% across a short-term gap (47.5 days in average), and reaching 18.4\% under the longest cross-session interval (1572 days in average). Notably, Heartprint covers a substantially broader longitudinal span than the other evaluated datasets. As a result, its short-term interval is comparable in duration to the long-term settings of the other databases. This broader temporal coverage supports the inclusion of Heartprint's intermediate short-term results in Figure~\ref{fig:long_term_eer_stability} to illustrate continuous, progressive template aging. The CYBHi dataset also shows a marked increase in error, from 12.1\% to 19.7\% over its long-term split. In contrast, the ECG-ID dataset exhibits greater resilience, with EER rising only from 6.6\% to 8.6\%. This relative stability is likely due to the highly controlled, resting conditions of clinical acquisitions, which exhibit less baseline drift than the more variable, off-the-person measurements in Heartprint and CYBHi. The results are aligned with recent findings in the literature. For instance, \cite{d2023advancing} demonstrates that identification accuracy degrades from $\approx$ 99\% to $\approx$ 70\% and from $\approx$ 99\% to $\approx$ 50\% for CYBHi and Heartprint datasets, respectively.

The findings underscore the necessity of evaluating biometric systems using protocols that enforce strict long-term temporal separation. In the absence of such protocols, reported performance tends to significantly overestimate the biometric system's actual capability and does not accurately represent the stability required for practical, long-term deployment.

\subsection{Efficiency of Multi-Session Enrollment}
\label{subsec:multi_session_rescue}

The observed inflation of verification error across both short-term and long-term evaluation regimes suggests that ECG morphology is not perfectly stationary over time. Over days, weeks, or months, factors such as changes in resting heart rate, gradual physiological adaptations, and minor variations in electrode placement can cause a subject's current ECG signal to deviate from the morphological baseline established during initial enrollment. As a result, biometric systems that rely on a single static enrollment recording, known as the Single-Shot (SS) protocol, are inherently susceptible to temporal drift. To determine whether this vulnerability can be reduced, a multi-session enrollment strategy based on the Leave-Last-Out (LLO) protocol is evaluated. In this approach, the user template is constructed by combining ECG beats from multiple historical sessions rather than depending solely on the initial recording. This method expands the representation of the user's physiological variability within the gallery template, enabling the biometric model to better accommodate natural temporal drift.

\begin{figure}[htbp]
    \centering
    \includegraphics[width=\linewidth]{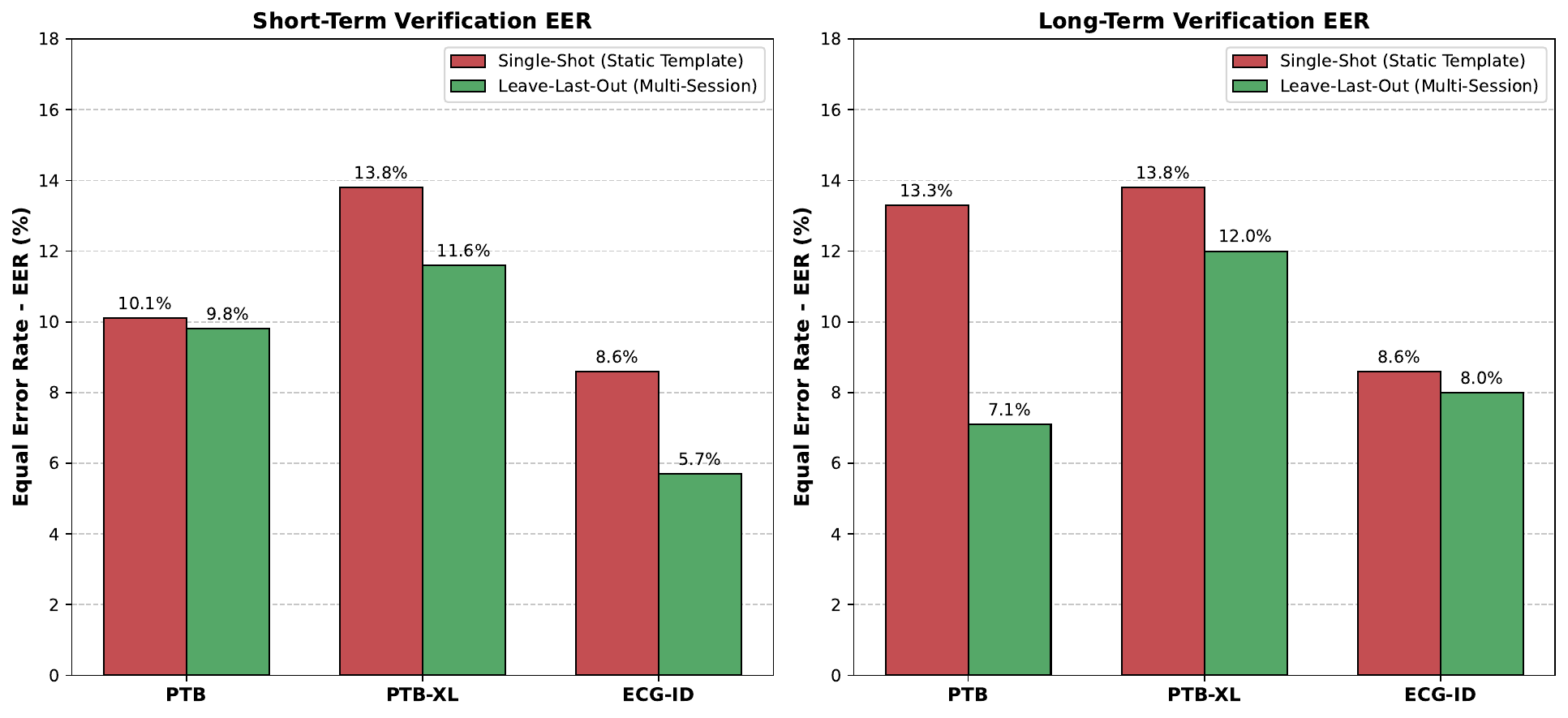}
    \caption{Verification Error Reduction via Multi-Session Enrollment.}
    \label{fig:llo_eer_reduction}
\end{figure}

To isolate the effect of template updating from population-scale factors, this study compares the 1:1 verification EER obtained using static Single-Shot enrollment and dynamic multi-session enrollment. As shown in Figure~\ref{fig:llo_eer_reduction}, incorporating multiple historical sessions consistently reduces verification error in both short-term and long-term evaluation regimes. In the short-term scenario, where the primary sources of variability are session-to-session electrode repositioning and day-to-day physiological fluctuations, multi-session enrollment yields noticeable improvements. For instance, on the ECG-ID dataset, the LLO protocol reduces the short-term EER from 8.6\% to 5.7\%, a 2.9 percentage point improvement. The effect is even more pronounced in long-term evaluations, where larger temporal gaps increase physiological variability. On the PTB dataset, the static Single-Shot template results in an EER of 13.3\% under the longest temporal separation, whereas the multi-session LLO strategy reduces the error to 7.1\%, a substantial improvement of 6.2 percentage points. Even on the relatively stable ECG-ID dataset, the long-term EER decreases from 8.6\% to 8.0\% with multi-session enrollment.

These observations highlight an important design consideration for practical ECG biometric systems. Static enrollment templates may gradually become outdated as the user's physiological signal evolves over time. In contrast, systems that periodically update templates with verified new recordings are better equipped to accommodate longitudinal variability, thereby supporting more consistent verification performance in long-term deployments.

\subsection{Generalizability and the Open-Set Challenge}
\label{subsec:generalizability_open_set}

A major limitation in the existing ECG biometrics literature is the predominant use of closed-set evaluation protocols. In a closed-set regime, the biometric system is assessed using subjects whose identities were previously available to the feature extractor during training. Although this approach demonstrates the model's ability to recognize known identities, practical deployment often requires open-set (subject-disjoint) operational capability. In such scenarios, the system must authenticate or reject previously unseen individuals without necessitating retraining of the underlying deep neural network. Despite the critical nature of this requirement, methodologies for robust open-set impostor rejection remain severely underexplored in the literature. Among the few exceptions addressing this gap is the work of \cite{d2023advancing}, which introduces a confidence-based rejection criterion to filter unauthorized identities.

\begin{figure}[htbp]
    \centering
    \includegraphics[width=\linewidth]{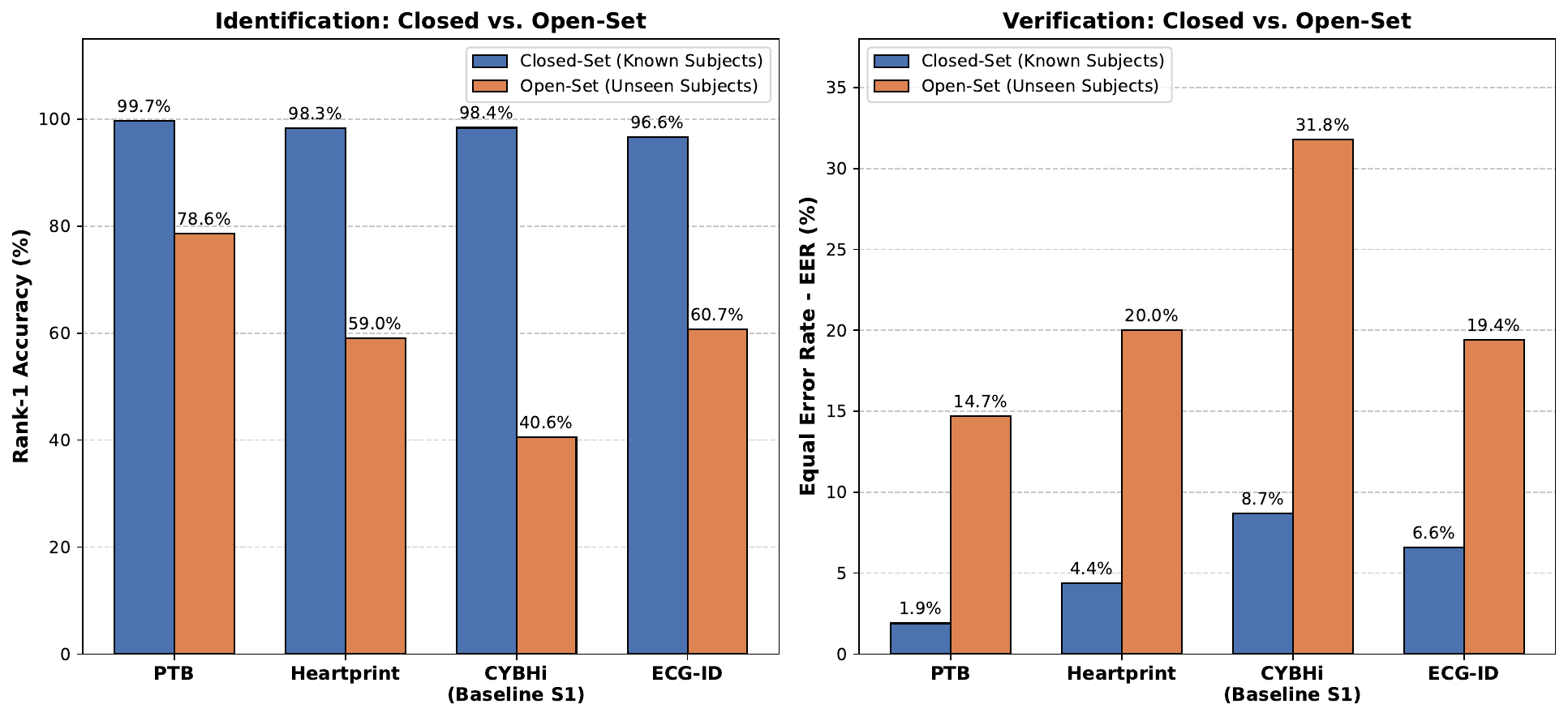}
    \caption{Model Generalizability: Closed-Set vs. Open-Set Performance.}
    \label{fig:generalizability_open_set}
\end{figure}

The generalizability of the learned feature space was evaluated by comparing the baseline performance of the model under both closed-set and open-set protocols. As shown in Figure~\ref{fig:generalizability_open_set}, the introduction of unseen subjects results in a decline in biometric performance across all evaluated datasets.
In the 1:N identification task (Rank-1), accuracy declines substantially. On the Heartprint dataset, Rank-1 performance decreases by 39.3 percentage points, while the CYBHi dataset experiences an even greater reduction of 57.8 percentage points, reaching only 40.6\%. These results suggest that the dense classification space established during training may be overfit to the known identities. When a previously unseen subject is introduced, the model fails to assign a distinct cluster for their morphology, resulting in frequent misclassification. The 1:1 verification task, measured by EER, demonstrates a similarly pronounced decline. The feature extractor does not learn universally discriminative physiological features, but instead relies on subject-specific features acquired during training. Consequently, the distances between genuine and impostor pairs for unseen subjects become indistinct. For example, the verification error on the clinical PTB dataset increases from 1.9\% to 14.7\%. More variable datasets, such as CYBHi, exhibit EER values exceeding 31.8\%.

A counterintuitive phenomenon sometimes emerges during evaluation. Rank-1 identification accuracy in the open-set regime may paradoxically exceed the closed-set baseline. This anomaly stems entirely from gallery size bias. In standard dataset partitioning, the open-set testing cohort is typically a much smaller, reserved subset of the total population, whereas the closed-set training cohort is larger. Since Rank-1 evaluates a 1:$N$ competition, a much smaller open-set gallery eliminates many potential distractors from the search space. As a result, the likelihood that a degraded or unfamiliar probe will inadvertently match an impostor decreases, which artificially inflates Rank-1 accuracy. However, referencing the 1:1 verification metrics resolves this paradox. Despite the inflated identification scores, the EER consistently worsens in open-set regimes, demonstrating that the underlying feature representation degrades when evaluated on unseen subjects.

\subsection{Controlled Hyperparameter Ablation}
\label{subsec:controlled_ablation}

\begin{figure}[htbp]
    \centering
    \includegraphics[width=\linewidth]{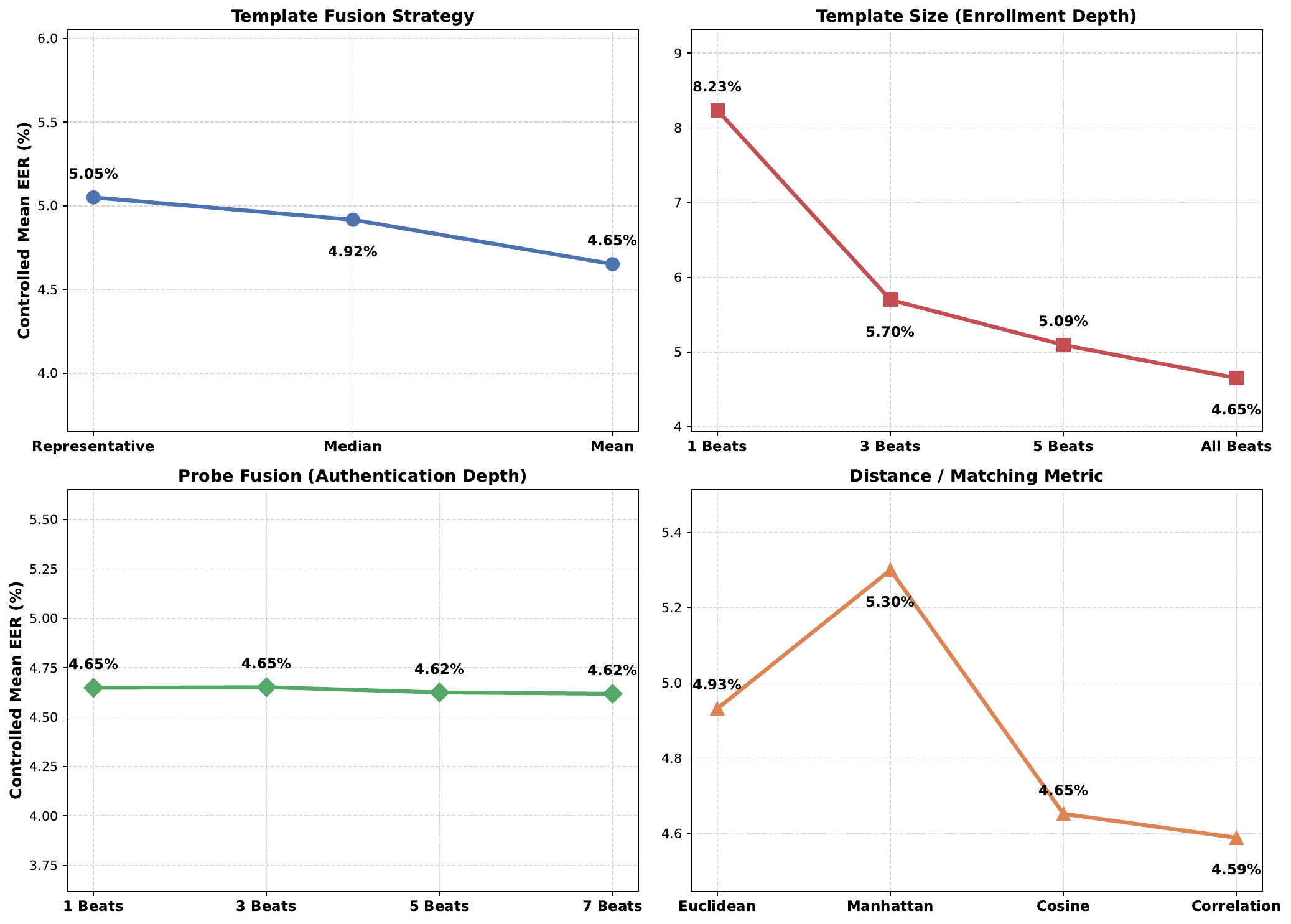}
    \caption{Controlled Hyperparameter Ablation (Single-Session Closed-Set).}
    \label{fig:controlled_hyperparameter_ablation}
\end{figure}

To isolate the marginal impact of individual system architecture choices, a controlled hyperparameter ablation study was conducted. In contrast to global averaging, which may be influenced by universally sub-optimal configurations, this ablation approach maintains the biometric system in a robust default state (Template Size = `All', Template Fusion = `Mean', Probe Fusion = 3, Metric = `Cosine'). By varying one parameter at a time across the single-session, closed-set evaluation regimes of all datasets, the analysis quantifies the primary performance drivers of the ECG biometric pipeline. As shown in Figure~\ref{fig:controlled_hyperparameter_ablation}, this approach identifies three critical system design constraints:

\text{1. Heavy Enrollment (Template Engineering):} The structure of the user's gallery template determines the maximum achievable system security. The top-right panel in Figure~\ref{fig:controlled_hyperparameter_ablation} illustrates a pronounced, monotonic reduction in verification error as enrollment depth increases. Enrollment with a single heartbeat results in a system that is highly susceptible to compromise (26.37\% Mean EER). In contrast, increasing the template size to three or five beats significantly enhances stability, with the lowest error (19.01\%) observed when all available beats from the enrollment window are utilized. Additionally, as depicted in the top-left panel, mathematically averaging these beats into a single continuous representation (Mean Fusion, 19.01\% EER) outperforms the selection of a single real heartbeat as a medoid (Representative Fusion, 24.08\% EER). These findings indicate that mean fusion effectively mitigates transient noise and establishes a robust morphological baseline. Our empirical conclusion that `Heavy Enrollment' establishes a robust morphological baseline is corroborated by recent literature. For instance, \cite{melzi2023ecg} independently validated this preprocessing logic by generating `template segments' via element-wise averaging of consecutive heartbeats. Their success in utilizing averaged templates to stabilize their deep Autoencoder aligns perfectly with our findings, confirming that multi-beat fusion is a universal prerequisite for reliable enrollment in deep learning pipelines.

\text{2. Lightweight Authentication (Probe Fusion):} In contrast to the substantial enrollment requirement, the authentication phase necessitates minimal temporal depth. The bottom-left panel demonstrates a consistently flat trendline for probe fusion. Once a robust, mean-fused gallery template is established, authenticating by fusing 3, 5, or 7 probe beats results in only a marginal improvement, reducing the EER by approximately 0.4 percentage points compared to a single beat. These findings support the feasibility of an efficient wearable deployment strategy: the system can acquire a dense, multi-beat recording during initial user setup and subsequently perform continuous background authentication using efficient, single-beat inference windows.

\text{3. Angular Distance Metrics:} Analysis of the matching functions (bottom-right panel) confirms the geometric characteristics of the learned feature space. Standard Euclidean distance, which depends on absolute spatial magnitudes, yields the poorest performance (20.69\% EER). In contrast, angle and shape-based metrics, specifically the Pearson Correlation Coefficient (18.91\%) and Cosine Similarity (19.01\%), provide the highest structural separability. These results suggest that identity-specific ECG morphology is primarily encoded in the relational directionality of the deep feature vectors rather than their absolute scale.

\subsection{Architectural Agnosticism: Cross-Model Validation}
\label{subsec:architectural_agnosticism}

A potential critique of large-scale biometric benchmarking is that identified vulnerabilities might simply be isolated artifacts of the specific neural network topology evaluated. To ensure the integrity of our insights and suggest that these limitations are intrinsic to current supervised paradigms, we extended our evaluation across various models. 

\begin{figure}[htbp]
    \centering
    \includegraphics[width=\linewidth]{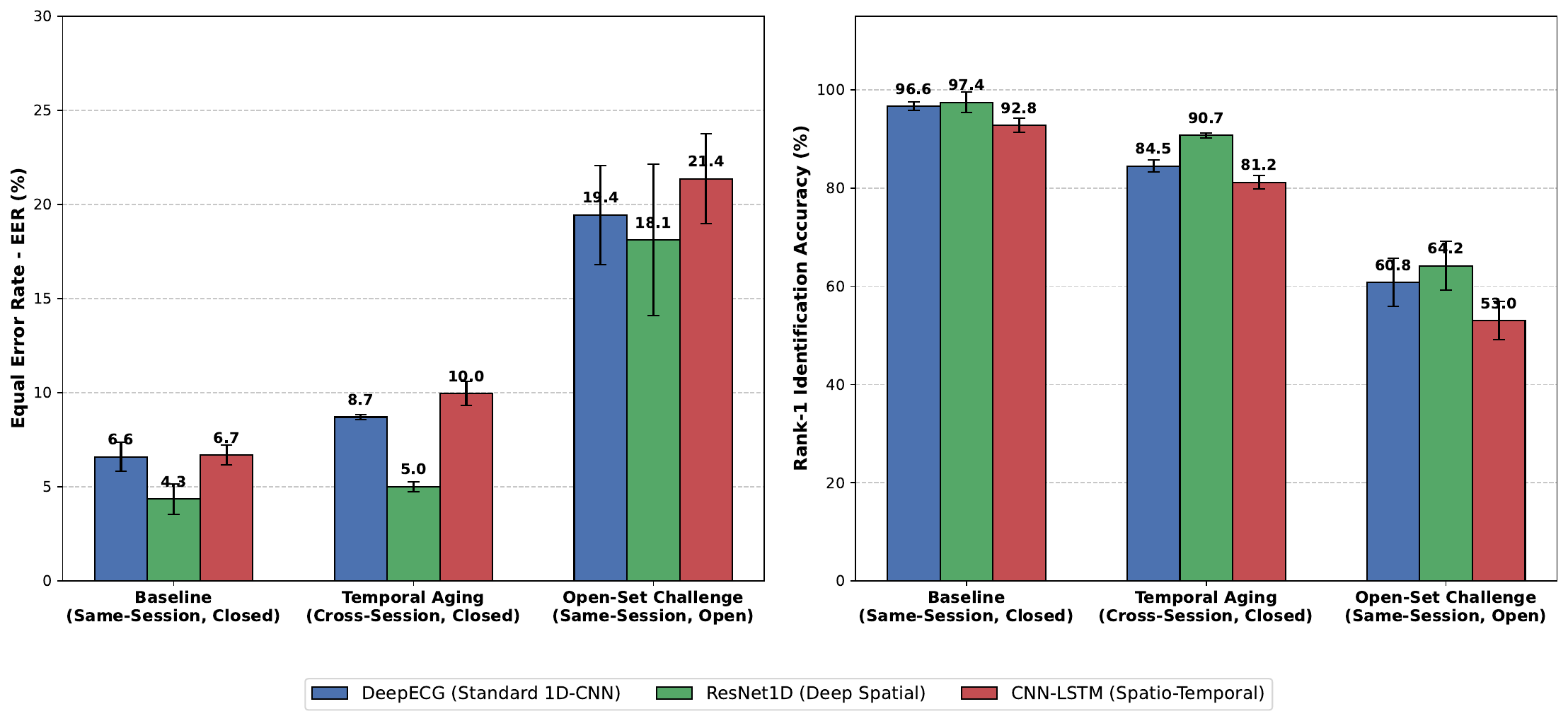}
    \caption{Architectural Agnosticism.}
    \label{fig:architectural_agnosticism}
\end{figure}

In this ablation, we compared the baseline 1D-CNN (DeepECG) against a significantly deeper convolutional architecture (ResNet1D) and a hybrid spatio-temporal network (CNN-LSTM). To ensure strict comparability, all models were evaluated using an identical hyperparameter configuration on the ECG-ID dataset across three progressive scenarios: an optimistic baseline (same-session, closed-set), a temporally aged regime (cross-session, closed-set), and an unseen identity regime (same-session, open-set). As illustrated in Figure~\ref{fig:architectural_agnosticism}, while absolute performance varies, the relative degradation trends remain consistent across architectures.

Under the optimistic baseline conditions, all architectures achieve highly competitive verification error rates, ranging from 4.35\% EER for ResNet1D to 6.68\% for the CNN-LSTM, confirming that diverse network topologies are fully capable of memorizing familiar, temporally adjacent ECG morphologies. However, the introduction of realistic evaluation constraints induces performance degradation across all architectures. When evaluating cross-session temporal separation, every model exhibits a drop in performance, with Rank-1 accuracies degrading by 6 to 15 percentage points. When evaluating an open-set scenario, all three architectures suffer a significant collapse in performance. The verification EER of the ResNet1D model more than quadruples to 18.12\%, while the CNN-LSTM and DeepECG models experience a similar degradation, reaching an EER of approximately 20\%. Notably, this systemic degradation extends even to attention-based paradigms, as evidenced by the long-term evaluation drops reported for Vision Transformers in \cite{d2023advancing}.

These cross-model results support that the vulnerabilities exposed by \textit{ECG-biometrics-bench} are likely intrinsic to current supervised feature-learning paradigms. The inability to generalize physiological traits across time and unseen populations is a modality-wide challenge that is unlikely to be resolved solely by upgrading the neural backbone.

\section{Limitations and Future Work}
\label{sec:limitations_future_work}

Although this study establishes a comprehensive framework for reproducible ECG biometric evaluation, several limitations remain that warrant further investigation.

Primarily due to the significant computational demands of executing over 6,000 unique hyperparameter grid-search configurations across multiple large-scale datasets, including PTB-XL, the most exhaustive analyses were conducted using the DeepECG (1D-CNN) architecture. Although subsequent multi-model validation demonstrates that key vulnerabilities, such as temporal aging and open-set challenge, persist across various architectures, future research can extend exhaustive hyperparameter sweeps to these alternative topologies to identify architecture-specific optimization strategies.

Second, open-set evaluation reveals a critical and unresolved vulnerability in modern ECG biometrics: current deep convolutional networks are highly susceptible to memorizing identity-specific noise rather than extracting universally generalizable physiological traits. When applied to previously unseen subjects, verification error rates increase substantially. Therefore, future research should move beyond standard classification paradigms and prioritize advanced metric learning and subject-disjoint training regimes that explicitly promote open-set generalizability.

Third, although the multi-session enrollment strategy effectively mitigates the effects of long-term temporal aging, the current approach depends on discrete, static historical fusions. Practical wearable deployments will necessitate the development of continuous, adaptive template-updating algorithms. Future research should explore dynamic weighting functions that gradually remove outdated morphological features while incorporating newly verified, high-quality ECG beats obtained through passive background monitoring.

Another limitation of the current study is the exclusive focus on deep learning-based feature extraction. While the results consistently demonstrate vulnerabilities such as temporal aging and poor cross-session generalization, it remains important to verify whether these effects are inherent to the ECG modality or primarily induced by representation learning biases. Prior studies based on conventional fiducial and non-fiducial feature engineering approaches have also reported performance degradation under cross-session and long-term evaluation protocols~\cite{odinaka2012ecg, pinto2018evolution}, suggesting that temporal variability is a modality-level challenge rather than a model-specific artifact. Nevertheless, a direct, controlled comparison within a unified framework is currently lacking. Future work will extend \textit{ECG-biometrics-bench} to incorporate classical feature extraction pipelines, enabling a systematic evaluation of whether traditional methods exhibit similar robustness limitations under identical experimental conditions.

Finally, in this work, we use the term `open-set' to refer to subject-disjoint evaluation, where identities in the test set are not present during training. While this protocol is widely adopted in ECG biometrics literature, it differs from the classical definition of open-set recognition, which explicitly requires the system to reject unknown identities based on decision thresholds or novelty detection mechanisms. Some prior works define open-set recognition in this stricter sense, incorporating impostor rejection and metrics such as Detection and Identification Rate (DIR). We acknowledge this distinction and note that our current framework evaluates generalization to unseen identities rather than full open-set recognition. As part of future work, we plan to extend \textit{ECG-biometrics-bench} to include explicit open-set protocols with threshold-based rejection, as well as additional security-oriented evaluation scenarios such as presentation attack resilience and adversarial robustness.

\section{Conclusion}
\label{sec:conclusion}

The transition of ECG biometrics from theoretical research into viable, real-world applications requires a fundamental shift in how biometric models are evaluated. In this paper, we introduced \textit{ECG-biometrics-bench}, a unified, open-source framework designed to replace the fragmented and often misleading evaluation protocols that currently dominate the literature. By systematically evaluating diverse deep learning architectures across multiple datasets, we exposed the `Random Split Fallacy'. We demonstrated that evaluating models using randomized intra-session splits exploits transient temporal correlations, resulting in artificially inflated accuracies that immediately collapse under realistic long-term, cross-session conditions. Furthermore, our analysis revealed the inherent deceptiveness of Rank-1 identification as a comparative metric, proving that it is mathematically biased by gallery size and routinely contradicts the true structural robustness measured by 1:1 verification metrics (EER and TAR@FAR). Moreover, through a massive, globally averaged hyperparameter ablation, we identified that the optimal approach for ECG recognition relies on a `Heavy Enrollment, Lightweight Authentication' paradigm. By enforcing a dense, multi-beat mean fusion during initial setup, systems can establish highly resilient morphological baselines. Once enrolled, these systems can leverage angle-based distance metrics to securely and passively authenticate users using efficient, single-beat inference windows. Ultimately, we hope that the \textit{ECG-biometrics-bench} framework will serve as a foundation for more rigorous evaluation standards, guiding future research toward the practical challenges of open-set generalizability and longitudinal ECG recognition.

\section*{Software Availability}
To support reproducible research, the complete source code, installation and usage instructions, and pretrained models for \textit{ECG-biometrics-bench} will be made publicly available on GitHub upon the acceptance of this article.


\bibliographystyle{IEEEtran}
\bibliography{references}  

@article{pinto2018evolution,
  title={Evolution, current challenges, and future possibilities in ECG biometrics},
  author={Pinto, Jo{\~a}o Ribeiro and Cardoso, Jaime S and Louren{\c{c}}o, Andr{\'e}},
  journal={Ieee Access},
  volume={6},
  pages={34746--34776},
  year={2018},
  publisher={IEEE}
}

@article{merone2017ecg,
  title={ECG databases for biometric systems: A systematic review},
  author={Merone, Mario and Soda, Paolo and Sansone, Mario and Sansone, Carlo},
  journal={Expert Systems with Applications},
  volume={67},
  pages={189--202},
  year={2017},
  publisher={Elsevier}
}

@article{melzi2023ecg,
  title={Ecg biometric recognition: Review, system proposal, and benchmark evaluation},
  author={Melzi, Pietro and Tolosana, Ruben and Vera-Rodriguez, Ruben},
  journal={IEEE Access},
  volume={11},
  pages={15555--15566},
  year={2023},
  publisher={IEEE}
}

@article{mesinovic2025survbench,
  title={SurvBench: A Standardised Preprocessing Pipeline for Multi-Modal Electronic Health Record Survival Analysis},
  author={Mesinovic, Munib and Zhu, Tingting},
  journal={arXiv preprint arXiv:2511.11935},
  year={2025}
}

@article{vest2018open,
  title={An open source benchmarked toolbox for cardiovascular waveform and interval analysis},
  author={Vest, Adriana N and Da Poian, Giulia and Li, Qiao and Liu, Chengyu and Nemati, Shamim and Shah, Amit J and Clifford, Gari D},
  journal={Physiological measurement},
  volume={39},
  number={10},
  pages={105004},
  year={2018},
  publisher={IOP Publishing}
}

@article{odinaka2012ecg,
  title={ECG biometric recognition: A comparative analysis},
  author={Odinaka, Ikenna and Lai, Po-Hsiang and Kaplan, Alan D and O'Sullivan, Joseph A and Sirevaag, Erik J and Rohrbaugh, John W},
  journal={IEEE Transactions on Information Forensics and Security},
  volume={7},
  number={6},
  pages={1812--1824},
  year={2012},
  publisher={IEEE}
}

@phdthesis{lugovaya2005biometric,
  author = {Lugovaya, T. S.},
  title = {Biometric human identification based on electrocardiogram},
  school = {Faculty of Computing Technologies and Informatics, Electrotechnical University "LETI", Saint-Petersburg, Russian Federation},
  year = {2005},
  month = {June}
}

@article{physionet2000,
  author = {Goldberger, Ary L. and Amaral, Luis A. N. and Glass, Leon and Hausdorff, Jeffrey M. and Ivanov, Plamen Ch. and Mark, Roger G. and Mietus, Joseph E. and Moody, George B. and Peng, Chung-Kang and Stanley, H. Eugene},
  title = {PhysioBank, PhysioToolkit, and PhysioNet: Components of a new research resource for complex physiologic signals},
  journal = {Circulation},
  year = {2000},
  volume = {101},
  number = {23},
  pages = {e215--e220}
}

@article{islam2022heartprint,
  title={Heartprint: A dataset of multisession ECG signal with long interval captured from fingers for biometric recognition},
  author={Islam, Md Saiful and Alhichri, Haikel and Bazi, Yakoub and Ammour, Nassim and Alajlan, Naif and Jomaa, Rami M},
  journal={Data},
  volume={7},
  number={10},
  pages={141},
  year={2022},
  publisher={MDPI}
}

@article{da2014check,
  title={Check Your Biosignals Here: A new dataset for off-the-person ECG biometrics},
  author={Da Silva, Hugo Pl{\'a}cido and Louren{\c{c}}o, Andr{\'e} and Fred, Ana and Raposo, Nuno and Aires-de-Sousa, Marta},
  journal={Computer methods and programs in biomedicine},
  volume={113},
  number={2},
  pages={503--514},
  year={2014},
  publisher={Elsevier}
}

@article{moody2001impact,
  title={The impact of the MIT-BIH arrhythmia database},
  author={Moody, George B and Mark, Roger G},
  journal={IEEE engineering in medicine and biology magazine},
  volume={20},
  number={3},
  pages={45--50},
  year={2001},
  publisher={IEEE}
}

@article{wagner2020ptb,
  title={PTB-XL, a large publicly available electrocardiography dataset},
  author={Wagner, Patrick and Strodthoff, Nils and Bousseljot, Ralf-Dieter and Kreiseler, Dieter and Lunze, Fatima I and Samek, Wojciech and Schaeffter, Tobias},
  journal={Scientific data},
  volume={7},
  number={1},
  pages={1--15},
  year={2020},
  publisher={Nature Publishing Group}
}

@article{labati2019deep,
  title={Deep-ECG: Convolutional neural networks for ECG biometric recognition},
  author={Labati, Ruggero Donida and Mu{\~n}oz, Enrique and Piuri, Vincenzo and Sassi, Roberto and Scotti, Fabio},
  journal={Pattern Recognition Letters},
  volume={126},
  pages={78--85},
  year={2019},
  publisher={Elsevier}
}

@article{zehir2024empirical,
  title={Empirical mode decomposition-based biometric identification using GRU and LSTM deep neural networks on ECG signals},
  author={Zehir, Hatem and Hafs, Toufik and Daas, Sara},
  journal={Evolving Systems},
  volume={15},
  number={6},
  pages={2193--2209},
  year={2024},
  publisher={Springer}
}

@article{sasikala2010identification,
  title={Identification of individuals using electrocardiogram},
  author={Sasikala, P and Wahidabanu, RSD},
  journal={International journal of computer science and network security},
  volume={10},
  number={12},
  pages={147--153},
  year={2010}
}

@article{belo2020ecg,
  title={ECG biometrics using deep learning and relative score threshold classification},
  author={Belo, David and Bento, Nuno and Silva, Hugo and Fred, Ana and Gamboa, Hugo},
  journal={Sensors},
  volume={20},
  number={15},
  pages={4078},
  year={2020},
  publisher={MDPI}
}

@article{pinto2017towards,
  title={Towards a continuous biometric system based on ECG signals acquired on the steering wheel},
  author={Pinto, Jo{\~a}o Ribeiro and Cardoso, Jaime S and Louren{\c{c}}o, Andr{\'e} and Carreiras, Carlos},
  journal={Sensors},
  volume={17},
  number={10},
  pages={2228},
  year={2017},
  publisher={MDPI}
}

@article{hejazi2016ecg,
  title={ECG biometric authentication based on non-fiducial approach using kernel methods},
  author={Hejazi, Maryamsadat and Al-Haddad, Syed Abdul Rahman and Singh, Yashwant Prasad and Hashim, Shaiful Jahari and Aziz, Ahmad Fazli Abdul},
  journal={Digital Signal Processing},
  volume={52},
  pages={72--86},
  year={2016},
  publisher={Elsevier}
}

@article{el2022wavelet,
  title={A wavelet-based capsule neural network for ECG biometric identification},
  author={El Boujnouni, Imane and Zili, Hassan and Tali, Abdelhak and Tali, Tarik and Laaziz, Yassin},
  journal={Biomedical Signal Processing and Control},
  volume={76},
  pages={103692},
  year={2022},
  publisher={Elsevier}
}

@article{wang2022ecg,
  title={An ECG signal denoising method using conditional generative adversarial net},
  author={Wang, Xiaoyu and Chen, Bingchu and Zeng, Ming and Wang, Yuli and Liu, Hui and Liu, Ruixia and Tian, Lan and Lu, Xiaoshan},
  journal={IEEE Journal of Biomedical and Health Informatics},
  volume={26},
  number={7},
  pages={2929--2940},
  year={2022},
  publisher={IEEE}
}

@article{al2024person,
  title={Person identification with arrhythmic ECG signals using deep convolution neural network},
  author={Al-Jibreen, Awabed and Al-Ahmadi, Saad and Islam, Saiful and Artoli, Abdel Momin},
  journal={Scientific Reports},
  volume={14},
  number={1},
  pages={4431},
  year={2024},
  publisher={Nature Publishing Group UK London}
}

@article{alduwaile2021using,
  title={Using convolutional neural network and a single heartbeat for ECG biometric recognition},
  author={AlDuwaile, Dalal A and Islam, Md Saiful},
  journal={Entropy},
  volume={23},
  number={6},
  pages={733},
  year={2021},
  publisher={MDPI}
}

@article{jyotishi2021ecg,
  title={An ECG biometric system using hierarchical LSTM with attention mechanism},
  author={Jyotishi, Debasish and Dandapat, Samarendra},
  journal={IEEE Sensors Journal},
  volume={22},
  number={6},
  pages={6052--6061},
  year={2021},
  publisher={IEEE}
}

@article{pan1985real,
  title={A real-time QRS detection algorithm},
  author={Pan, Jiapu and Tompkins, Willis J},
  journal={IEEE transactions on biomedical engineering},
  number={3},
  pages={230--236},
  year={1985},
  publisher={IEEE}
}

@inproceedings{hamilton2002open,
  title={Open source ECG analysis},
  author={Hamilton, Pat},
  booktitle={Computers in cardiology},
  pages={101--104},
  year={2002},
  organization={IEEE}
}

@article{christov2004real,
  title={Real time electrocardiogram QRS detection using combined adaptive threshold},
  author={Christov, Ivaylo I},
  journal={Biomedical engineering online},
  volume={3},
  pages={1--9},
  year={2004},
  publisher={Springer}
}

@article{makowski2021neurokit2,
  title={NeuroKit2: A Python toolbox for neurophysiological signal processing},
  author={Makowski, Dominique and Pham, Tam and Lau, Zen J and Brammer, Jan C and Lespinasse, Fran{\c{c}}ois and Pham, Hung and Sch{\"o}lzel, Christopher and Chen, SH Annabel},
  journal={Behavior research methods},
  pages={1--8},
  year={2021},
  publisher={Springer}
}

@article{labati2023multicardionet,
  title={MultiCardioNet: Interoperability between ECG and PPG biometrics},
  author={Labati, Ruggero Donida and Piuri, Vincenzo and Rundo, Francesco and Scotti, Fabio},
  journal={Pattern Recognition Letters},
  volume={175},
  pages={1--7},
  year={2023},
  publisher={Elsevier}
}

@article{ammour2023deep,
  title={Deep contrastive learning-based model for ECG biometrics},
  author={Ammour, Nassim and Jomaa, Rami M and Islam, Md Saiful and Bazi, Yakoub and Alhichri, Haikel and Alajlan, Naif},
  journal={Applied Sciences},
  volume={13},
  number={5},
  pages={3070},
  year={2023},
  publisher={MDPI}
}

@article{wang2024ecg,
  title={ECG biometric authentication using self-supervised learning for IoT edge sensors},
  author={Wang, Guoxin and Shanker, Shreejith and Nag, Avishek and Lian, Yong and John, Deepu},
  journal={IEEE Journal of Biomedical and Health Informatics},
  year={2024},
  publisher={IEEE}
}

@article{hammad2019parallel,
  title={Parallel score fusion of ECG and fingerprint for human authentication based on convolution neural network},
  author={Hammad, Mohamed and Wang, Kuanquan},
  journal={Computers \& Security},
  volume={81},
  pages={107--122},
  year={2019},
  publisher={Elsevier}
}

@article{pereira2023biometric,
  title={Biometric recognition: A systematic review on electrocardiogram data acquisition methods},
  author={Pereira, Teresa MC and Concei{\c{c}}{\~a}o, Raquel C and Sencadas, Vitor and Sebasti{\~a}o, Raquel},
  journal={Sensors},
  volume={23},
  number={3},
  pages={1507},
  year={2023},
  publisher={MDPI}
}

@article{schijvenaars2008intraindividual,
  title={Intraindividual variability in electrocardiograms},
  author={Schijvenaars, Bob JA and van Herpen, Gerard and Kors, Jan A},
  journal={Journal of Electrocardiology},
  volume={41},
  number={3},
  pages={190--196},
  year={2008},
  publisher={Elsevier}
}

@article{rai2025lightweight,
  title={Lightweight MobileNetV1+ GRU for ECG Biometric Authentication: Federated and Adversarial Evaluation},
  author={Rai, Dilli Hang and Kafley, Sabin},
  journal={arXiv preprint arXiv:2509.20382},
  year={2025}
}

@article{d2023advancing,
  title={Advancing ecg biometrics through vision transformers: A confidence-driven approach},
  author={D’angelis, Onorato and Bacco, Luca and Vollero, Luca and Merone, Mario},
  journal={IEEE Access},
  volume={11},
  pages={140710--140721},
  year={2023},
  publisher={IEEE}
}

@article{pereira2013novel,
  title={Novel fiducial and non-fiducial approaches to electrocardiogram-based biometric systems},
  author={Pereira Coutinho, David and Silva, Hugo and Gamboa, Hugo and Fred, Ana and Figueiredo, Mario},
  journal={IET biometrics},
  volume={2},
  number={2},
  pages={64--75},
  year={2013},
  publisher={Wiley Online Library}
}

@article{ciocoiu2017comparative,
  title={Comparative analysis of bag-of-words models for ECG-based biometrics},
  author={Ciocoiu, Iulian B},
  journal={IET Biometrics},
  volume={6},
  number={6},
  pages={495--502},
  year={2017},
  publisher={Wiley Online Library}
}

@inproceedings{zhao2011ecg,
  title={ECG identification based on matching pursuit},
  author={Zhao, Zhidong and Yang, Lei},
  booktitle={2011 4th International conference on biomedical engineering and informatics (BMEI)},
  volume={2},
  pages={721--724},
  year={2011},
  organization={IEEE}
}

@article{tantawi2015wavelet,
  title={A wavelet feature extraction method for electrocardiogram (ECG)-based biometric recognition},
  author={Tantawi, Manal M and Revett, Kenneth and Salem, Abdel-Badeeh and Tolba, Mohamed F},
  journal={Signal, Image and Video Processing},
  volume={9},
  number={6},
  pages={1271--1280},
  year={2015},
  publisher={Springer}
}

@article{irvine2008eigenpulse,
  title={eigenPulse: Robust human identification from cardiovascular function},
  author={Irvine, John M and Israel, Steven A and Scruggs, W Todd and Worek, William J},
  journal={Pattern Recognition},
  volume={41},
  number={11},
  pages={3427--3435},
  year={2008},
  publisher={Elsevier}
}

@article{camara2018real,
  title={Real-time electrocardiogram streams for continuous authentication},
  author={Camara, Carmen and Peris-Lopez, Pedro and Gonzalez-Manzano, Lorena and Tapiador, Juan},
  journal={Applied Soft Computing},
  volume={68},
  pages={784--794},
  year={2018},
  publisher={Elsevier}
}

@inproceedings{venkatesh2010human,
  title={Human electrocardiogram for biometrics using DTW and FLDA},
  author={Venkatesh, N and Jayaraman, Srinivasan},
  booktitle={2010 20th International Conference on Pattern Recognition},
  pages={3838--3841},
  year={2010},
  organization={IEEE}
}

@article{zhang2017heartid,
  title={HeartID: A multiresolution convolutional neural network for ECG-based biometric human identification in smart health applications},
  author={Zhang, Qingxue and Zhou, Dian and Zeng, Xuan},
  journal={Ieee Access},
  volume={5},
  pages={11805--11816},
  year={2017},
  publisher={IEEE}
}

@inproceedings{jahiruzzaman2015ecg,
  title={ECG based biometric human identification using chaotic encryption},
  author={Jahiruzzaman, Md and Hossain, ABM Aowlad},
  booktitle={2015 International Conference on Electrical Engineering and Information Communication Technology (ICEEICT)},
  pages={1--5},
  year={2015},
  organization={IEEE}
}

@article{bousseljot1995nutzung,
  title={Nutzung der EKG-Signaldatenbank CARDIODAT der PTB {\"u}ber das Internet},
  author={Bousseljot, Ralf and Kreiseler, Dieter and Schnabel, Allard},
  year={1995},
  publisher={Walter de Gruyter, Berlin/New York Berlin, New York}
}

@article{goldberger2000physiobank,
  title={PhysioBank, PhysioToolkit, and PhysioNet: components of a new research resource for complex physiologic signals},
  author={Goldberger, Ary L and Amaral, Luis AN and Glass, Leon and Hausdorff, Jeffrey M and Ivanov, Plamen Ch and Mark, Roger G and Mietus, Joseph E and Moody, George B and Peng, Chung-Kang and Stanley, H Eugene},
  journal={circulation},
  volume={101},
  number={23},
  pages={e215--e220},
  year={2000},
  publisher={Lippincott Williams \& Wilkins}
}

\end{document}